\documentclass[twoside,twocolumn]{article}
\usepackage{PRIMEarxiv}
\usepackage{listings}

\usepackage{soul}
\usepackage[dvipsnames]{xcolor}
\usepackage{graphicx}
\usepackage{multirow}
\usepackage{amsmath,amssymb,amsfonts}
\usepackage{physics}
\usepackage{booktabs}
\usepackage{siunitx}
\usepackage[utf8]{inputenc} 
\usepackage[T1]{fontenc}    
\usepackage{hyperref}       
\usepackage{url}            
\usepackage{nicefrac}       
\usepackage{microtype}      
\usepackage{fancyhdr}       
\usepackage{pifont}
\usepackage{mathalpha}
\usepackage{cancel}
\usepackage{booktabs}

\definecolor{bg}{rgb}{0.95,0.95,0.95}
\definecolor{commentgreen}{rgb}{0,0.6,0}
\definecolor{stringred}{rgb}{0.58,0,0.82}

\lstdefinestyle{python}{
  language=Python,
  basicstyle=\small\ttfamily,
  backgroundcolor=\color{bg},
  keywordstyle=\color{blue},
  commentstyle=\color{commentgreen},
  stringstyle=\color{stringred},
  breaklines=true,
  showstringspaces=false,
  numbers=none,
  frame=none,
  xleftmargin=0pt,
  xrightmargin=0pt,
  aboveskip=0.5em,
  belowskip=0.5em
}

\newcommand{\pyinline}[1]{\lstinline[style=python, basicstyle=\small\ttfamily]|#1|}

\newcommand{\bs}[1]{\vb*{#1}}

\DeclareMathOperator*{\argmin}{arg\,min}

\usepackage[colorinlistoftodos,prependcaption,textsize=tiny,textcolor=white]{todonotes}
\usepackage{marginnote}   
\usepackage{etoolbox}     
\usepackage{ifthen}       
\newtoggle{ama@left}
\togglefalse{ama@left}

\usepackage{algorithm, algorithmicx, algpseudocode}
\newcommand*{\algrule}[1][\algorithmicindent]{\hspace*{.5em}\vrule\vrule width 0pt height \baselineskip depth .25\baselineskip\hspace*{\dimexpr#1-.5em}}
\makeatletter
\newcount\ALG@printindent@tempcnta
\def\ALG@printindent{%
    \ifnum \theALG@nested>0
    \ifx\ALG@text\ALG@x@notext
    \else
    \unskip
    \ALG@printindent@tempcnta=1
    \loop
    \algrule[\csname ALG@ind@\the\ALG@printindent@tempcnta\endcsname]%
    \advance \ALG@printindent@tempcnta 1
    \ifnum \ALG@printindent@tempcnta<\numexpr\theALG@nested+1\relax
    \repeat
    \fi
    \fi
}%
\usepackage{etoolbox}
\patchcmd{\ALG@doentity}{\noindent\hskip\ALG@tlm}{\ALG@printindent}{}{\errmessage{failed to patch}}
\makeatother
\AtBeginEnvironment{algorithmic}{\lineskip0pt}
\usepackage[scaled]{beramono}
\usepackage[T1]{fontenc}
\algtext*{EndWhile}
\algtext*{EndFor}
\algtext*{EndIf}
\algdef{SE}[DOWHILE]{Do}{doWhile}{\algorithmicdo}[1]{\algorithmicwhile\ #1}%
\algnewcommand\algorithmicto{\textbf{to}}
\algrenewtext{While}{\textbf{while}\ }
\algrenewtext{Procedure}[2]{\textbf{function}\ \textproc{#1}\ifthenelse{\equal{#2}{}}{}{(#2)}}

\algrenewcommand\algorithmicrequire{\textbf{Input:}}
\algrenewcommand\algorithmicensure{\textbf{Output:}}

\usepackage{eqparbox}
\algrenewcommand{\algorithmiccomment}[1]{\hfill\eqparbox{COMMENT}{\color{gray} \it-- #1}}

\title{Bilevel Optimization of Topology and Hyperparameters (BOTH)}

\author{
	Suryanarayanan Manoj Sanu, Alejandro M.~Arag\'{o}n \\
	Faculty of Mechanical Engineering\\
	Delft university of technology \\
	Delft\\
	The Netherlands \\
	\texttt{\{s.manojsanu, a.m.aragon\}@tudelft.nl} \\
	\And
	Miguel A.~Bessa \\
	School of engineering \\
	Brown University\\
	Providence\\
	United States of America \\
	\texttt{miguel\_bessa@brown.edu} \\
}

\begin{document}
\twocolumn[
\begin{@twocolumnfalse}
\maketitle

\begin{abstract}
Topology optimization (TO) represents a significant step towards automating the design process: given a working simulation, TO can produce a viable prototype at the press of a button by differentiating the simulation and iteratively improving the design. In practice, however, TO is riddled with ``magic numbers''---hyperparameters whose tuning significantly affects the outcome. Finding the right values typically requires not only deep problem-specific knowledge but also extensive trial-and-error. While practitioners can use surrogate-assisted hyperparameter optimization as an alternative, this approach requires strictly limiting the number of hyperparameters through careful problem formulation.
Here, we propose differentiating TO itself using automatic differentiation. This yields ``hypergradients'' that allow us to tune these hyperparameters in tandem with the primary optimization. We show that evaluating just one or two steps of TO is sufficiently informative and that the method scales favorably to thousands of hyperparameters at an expense comparable to only a few standard TO runs. We demonstrate this approach on stress-constrained and compliance problems, with the latter utilizing a neural parameterization of the density field.
\end{abstract}

\keywords{hyperparameter tuning, automatic differentiation, topology optimization, stress constraints, bilevel programming}

\vspace{1.5em}
\end{@twocolumnfalse}
]

\section{Introduction}

Topology optimization (TO,~\cite{bendsoe2003topology}), particularly the density-based formulation~\cite{Bendse1989}, gives rise to a high-dimensional design space in which each pixel (or voxel) can independently influence the evolving topology. This design freedom has enabled TO to generate structures with performance and complexity that often exceed human intuition~\cite{Aage2017, zhu2016topology, wang2017grid, li2022innovative}. Yet, in practice, TO's apparent automation of the design process conceals a substantial layer of algorithmic complexity. The optimized design depends not only on the physical problem definition, but also on numerous \emph{hyperparameters}---externally chosen parameters that govern \emph{how} the optimization problem is solved rather than \emph{what} is being designed. Unlike the design variables, they are set by the user before the optimization begins and are not updated by the optimizer itself. Combined with the non-convexity of TO, this sensitivity means that 
obtaining good designs still relies heavily on expert intuition and 
manual trial-and-error rather than on principled selection.

In practical TO workflows, hyperparameters enter at nearly every 
stage of the algorithm. The material interpolation model (e.g., solid isotropic material with penalization (SIMP)) introduces interpolation exponents governing the drive toward 
discreteness. The Design parameterization requires choosing an initial state (either a density field or a level-set function) that can strongly bias or even dictate the final topology. This dependency is especially critical in level-set methods, where standard formulations cannot nucleate new holes. On top of this, nearly all TO workflows incorporate one 
or more filtering and projection techniques for controlling length scales, discreteness, 
and---in some formulations---manufacturability, which introduce additional 
hyperparameters~\cite{Wang2010, 
Langelaar2016}. The state and adjoint solvers then add preconditioners, convergence 
tolerances, and initialization strategies; these choices not only 
affect computational cost but can even change the final 
topology~\cite{Amir2024}. Problems with many constraints, such as stress-constrained formulations~\cite{Verbart2016, silva_stress, Senhora2020}, 
further require aggregation strategies (e.g., $p$-norm), which introduce aggregation exponents, relaxation 
parameters for singularity removal, and penalty or augmented Lagrangian 
parameters. The 
optimizer contributes its own set: \textit{method of moving asymptotes} (MMA, ~\cite{Svanberg1987}), for 
instance, introduces move limits and asymptote-adaptation parameters. 
Finally, recent learning-based components such as neural 
reparameterizations and learned surrogates~\cite{hoyer, Woldseth2022} 
add yet another layer, spanning network architecture, training, and 
regularization choices. 
Recent work~\cite{Ha2024} has shown that even for compliance minimization, nearly half of runs in a large hyperparameter sweep failed to progress, highlighting the sensitivity of TO to hyperparameter 
selection.

Historically, experts have relied on heuristic rules to set 
hyperparameters. In density-based compliance minimization, for instance, 
the SIMP penalty parameter is often fixed at a value of 
three~\cite{andreassen2011efficient}, although certain formulations 
benefit from starting at one---which yields a convex sub-problem---and increasing it according to a schedule, a 
procedure known as continuation~\cite{peter_sigmund_contin, 
deaton2014survey}. Projection filter parameters, which sharpen the 
density field toward a discrete design, are similarly ramped 
heuristically~\cite{Wang2010, Lazarov2016}. Yet despite widespread adoption, 
optimization under these schedules is fragile: it can fail to 
converge even on simple benchmark problems~\cite{Stolpe2001}, and the 
choice of when and how aggressively to increase parameters remains 
largely at the practitioner's discretion. In more challenging settings 
such as stress-constrained optimization, the situation deteriorates 
further---heuristic choices for relaxation and aggregation parameters 
often determine whether optimization succeeds or fails~\cite{liu2018current}. The underlying difficulty is that the 
appropriate hyperparameter values depend strongly on the problem 
formulation, mesh resolution, and objective, leaving heuristic rules 
without reliable transferability.

This need for systematic hyperparameter selection has prompted a growing body of research aimed at replacing ad hoc heuristics with principled, automated approaches. Within this paradigm, black-box optimization methods such as Bayesian optimization (BO) learn to predict the 
final design performance from a given hyperparameter configuration, 
enabling efficient search over the hyperparameter space. For instance, \cite{Lynch2019} proposed a two-stage approach in which a new problem's features are first matched against a dataset of prior TO runs to suggest an initial hyperparameter configuration based on similarity, which is then refined through BO. \cite{Jiang2020} trained an image-based classifier on prior TO runs to predict design feasibility, coupling it with particle swarm optimization to identify MMA hyperparameter configurations that reliably produce feasible designs in the context of Moving Morphable Components TO. In a related but distinct context, \cite{Bujny2023} applied polynomial regression within similarity-based TO, where the optimization additionally seeks designs that resemble a reference structure; their model predicts the energy scaling factor that achieves a target similarity level, bypassing the need to sweep over multiple TO runs. \cite{Bacciaglia2025} employed a neural network surrogate trained on pre-computed datasets alongside BO to tune hyperparameters for frequency optimization problems.

Despite their differences in scope and problem setting, all of these approaches share a common dependency: they require pre-collected datasets of TO runs, which can be expensive to generate. \cite{Ha2024} addressed this by casting hyperparameter tuning as a bilevel optimization problem, eliminating the need for any pre-collected dataset. In their formulation, the upper-level (UL) employs surrogate-based 
optimization---iteratively constructing and minimizing an 
interpolating surrogate of the UL objective---to propose hyperparameter 
configurations, while the lower-level (LL) solves the TO problem from scratch for each proposal. Across compliance, mechanism design, and buckling problems, this approach substantially reduced the rate of failed or poorly converged runs and showed that hyperparameter sensitivity becomes more pronounced in nonlinear and multiphysics settings~\cite{Ha2024}. However, all methods discussed thus far rely on derivative-free optimizers. These approaches face intrinsic scaling limits as the 
number of hyperparameters grows: constructing a reliable surrogate 
requires sample counts that grow rapidly with dimensionality, and 
each sample requires solving the full inner TO problem. This limitation also constrains the formulations themselves: TO works only with a handful of hyperparameters because no available tuning method could effectively handle larger hyperparameter spaces. Removing this limitation would open 
the door to richer parameterizations---spatially varying filters, 
element-wise penalization, and beyond---that have remained largely unexplored.

To overcome this limitation, we propose---for the first time in 
TO---a bilevel gradient-based framework that simultaneously optimizes 
the design variables and the hyperparameters (\emph{BOTH}). The framework draws on 
the use of \emph{hypergradients} in the machine learning (ML) community, 
where differentiation through optimization loops has enabled 
the tuning of millions of hyperparameters~\cite{lorraine2020optimizing,
shaban2019truncated, liu2018darts, rajeswaran2019meta}. Transferring this 
machinery to TO is not straightforward. Usual bilevel formulations in ML 
distinguish the two levels only through the data used---the same loss 
function evaluated on training and validation sets, respectively---and 
the cost of evaluating the LL loss is relatively inexpensive. TO, by contrast, demands for vast computational resources due 
to the nested physics simulation, while also having to satisfy constraints. We address these 
challenges through three explicit design choices: we construct distinct LL and UL objectives tailored to the engineering problem; we treat 
constraints through a quadratic penalty method; and we mitigate the 
computational burden and the well-known instabilities of 
hypergradient estimation by truncating the inner trajectory to only 
a few steps, beginning with only \textit{one or two}. We first verify the framework on standard bilevel benchmarks, where we identify the key challenges of hypergradient construction and show that 
the framework mitigates them. We then demonstrate that even a few steps 
of TO yield gradient information sufficient to drive the UL 
optimization effectively. The framework is then applied to TO 
problems of increasing difficulty: compliance minimization, including 
the hyperparameter-sensitive case of neural 
reparameterizations~\cite{Sanu2025}, and stress-constrained 
optimization. The result is a framework that enables simultaneous 
tuning of large hyperparameter spaces at a cost that scales favorably 
with their dimensionality, thereby enabling TO formulations that were
previously inaccessible.

\section{Bilevel optimization formulation}

Hyperparameter optimization~\cite{HPOinML, Bischl2023} is naturally cast as a bilevel optimization problem~\cite{Bard1998}, with the general formulation\footnote{Specifically, this is the \textit{optimistic} formulation, in which the LL solver is assumed to return the solution most favorable to the UL objective. This is consistent with standard practice in the machine learning literature~\cite{Sinha2025, bennett2008bilevel, lorraine2020optimizing} and is more scalable than the \textit{pessimistic} alternative~\cite{Liu2022}.}:
\begin{equation}
\label{eq:bilevel}
\begin{aligned}
& \min_{\bs{y}}
  &&F\! \left(\bs{y}, \bs{x}_\star(\bs{y})\right) \\
& \text{such that} \quad 
&& \bs{x}_\star= \argmin_{\bs{x}}
\quad f(\bs{x}; \bs{y})
\end{aligned}
\end{equation}
where $\boldsymbol{y} \in \mathcal{Y} = \{\boldsymbol{y} \in \mathbb{R}^m : \boldsymbol{y}_{\min} \leq \boldsymbol{y} \leq \boldsymbol{y}_{\max}\}$ are the UL variables corresponding to the hyperparameters to be tuned, and $\boldsymbol{x} \in \mathcal{X}(\boldsymbol{y}), \;
\mathcal{X}(\boldsymbol{y}) = \{ \boldsymbol{x} \in \mathbb{R}^n :
\boldsymbol{x}_{\min} \leq \boldsymbol{x} \leq \boldsymbol{x}_{\max}, \;
g(\boldsymbol{x}; \boldsymbol{y}) \leq \boldsymbol{0}, \;
h(\boldsymbol{x}; \boldsymbol{y}) = \boldsymbol{0} \}$ are the LL variables, i.e., the design variables of the original 
optimization problem. $F$ and $f$ are the UL and LL objective functions, respectively. The general inequality and equality constraints at the LL, respectively $g $ and $ 
h$, are handled via quadratic penalty or augmented 
Lagrangian methods, while the box constraints on $\boldsymbol{x}$ 
and $\boldsymbol{y}$ are enforced via projection or reparameterization, 
as discussed later.

Finding an optimized solution $\left(\bs{y}^\star, \bs{x}_\star\right)$ to Eqn.~\eqref{eq:bilevel} when the UL dimensionality is high requires a gradient-based approach~\cite{HPOinML}. We maintain the nested (or bilevel) formulation, in which 
the LL variables optimize only the LL objective
while the UL variables optimize only the UL objective, preserving the role separation between design variables and 
hyperparameters. UL updates are then performed using gradients computed through the LL optimization, a class of methods known as hypergradient methods~\cite{lorraine2020optimizing, pmlr-v80-franceschi18a}. The hypergradient---the total derivative of $F$ with respect to $\bs{y}$---is given by:

\begin{equation}
    \frac{dF}{d\bs{y}} = \frac{\partial F}{\partial \bs{y}} + \frac{\partial F}{\partial \bs{x}_\star}\frac{\partial \bs{x}_\star}{\partial \bs{y}},
\label{eqn:hpgrad}
\end{equation}
\noindent where the first term is the direct partial term (often zero unless the function $F$ depends explicitly on hyperparameters), and the second term is the gradient through the LL optimization, utilizing the \textit{best-response Jacobian}\footnote{The best-response Jacobian is structurally analogous to the sensitivity term in adjoint analysis in TO, where the derivative of the displacement field $\boldsymbol{u}$ with respect to the densities $\boldsymbol{\rho}$ is obtained implicitly: in both cases, the sensitivity of an implicitly defined quantity to upstream parameters follows from applying the IFT to a defining equation---LL optimality here, the equilibrium equation there~\cite{Sanu2026}.} (i.e., $\dfrac{\partial \bs{x}_\star}{\partial \bs{y}}$). Applying the Implicit Function Theorem (IFT)~\cite{Krantz2013} to the LL optimality condition for unconstrained optimization, i.e., the gradient of the LL objective with respect to the LL parameters vanish at the optimum (${\nabla}_{\bs{x}} f({\bs{x}_\star}; \bs{y}) = \boldsymbol{0}$), we obtain:
\begin{equation}\label{eqn:uncongrad}
    \left. \dfrac{\partial \bs{x}_\star}{\partial \bs{y}} = -\mathbf{H}_f^{-1}\dfrac{\partial^2 f}{\partial \bs{y} \partial \bs{x}} \right|_{\bs{x_\star}},
\end{equation}
where $\mathbf{H}_f = \left.\dfrac{\partial^2 f}{\partial \bs{x}^2} \right|_{\bs{x_\star}} \in \mathbb{R}^{n\times n}$ is the Hessian matrix of the LL objective function evaluated at a LL optimum~\cite{gould2021deep}. Each UL update requires solving an $n \times n$ linear system, for which a range of techniques have been developed 
to avoid forming $\mathbf{H}_f^{-1}$ explicitly~\cite{lorraine2020optimizing}.

\begin{figure*}
    \centering
    \includegraphics[width=1.0\linewidth]{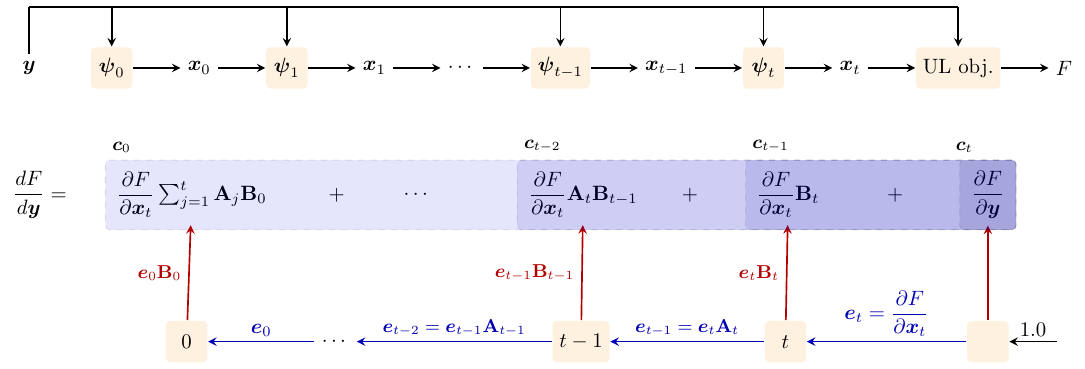}
    \caption{Reverse-mode automatic differentiation of the hypergradient. 
    \textbf{Top:} forward computational graph for the bilevel problem, 
    showing each optimizer step $\boldsymbol{{\psi}}_j$ as a node taking 
    both the previous iterate $\boldsymbol{x}_{j-1}$ and the UL variables 
    $\boldsymbol{y}$ as inputs, with the UL objective $F$ evaluated at 
    the final iterate $\boldsymbol{x}_t$. \textbf{Bottom:} reverse-mode backward 
    pass. A unit seed enters at $F$ and propagates backward through the 
    trajectory (blue arrows) via the recursion $\boldsymbol{e}_{j-1} = 
    \boldsymbol{e}_j \mathbf{A}_j$. At each step, a second branch (red 
    arrows) contributes $\boldsymbol{e}_j \mathbf{B}_j$ to the register, 
    where the hypergradient is accumulated as $\boldsymbol{c}_{j-1} = 
    \boldsymbol{c}_j + \boldsymbol{e}_j \mathbf{B}_j$. The final 
    hypergradient $dF / d\boldsymbol{y}$ is read off as $\boldsymbol{c}_0$ 
    at the leftmost endpoint of the register.}
    \label{fig:ad_unroll}
\end{figure*}

While Eqs.~\eqref{eqn:hpgrad}--\eqref{eqn:uncongrad} represent the exact hypergradient, this IFT-based formulation has several practical limitations: First, each UL update requires solving a large linear system whose matrix may be severely ill-conditioned. Second, the hypergradient is 
exposed to two distinct sources of error~\cite{shen2024memory}: the inexactness of the LL 
solution (since attaining true optimality for every hyperparameter configuration is 
computationally prohibitive for TO), 
and the convergence tolerance of the iterative linear solve, which amplifies with the condition number of the matrix. 
Third, the IFT framework formally requires the LL to be both strongly 
convex (so that $\mathbf{H}_f$ is invertible)~\cite{pmlr-v202-shen23c} and smooth~\cite{pedregosa16}. Finally, hyperparameters that only affect the LL optimization trajectory (and not LL loss landscape itself)---such as initializations, tolerances, relaxations, and optimizer hyperparameters---cannot be tuned under the IFT formulation, since the optimality criterion does not depend on any of them~\cite{rajeswaran2019meta, lorraine2020optimizing}.

For these reasons, we adopt an automatic differentiation (AD) approach that estimates  hypergradients by explicitly differentiating through the LL optimization trajectory. We relax the requirement of LL optimality by replacing $\bs{x}_\star$ with a finite-step estimate $\bs{x}_t$, obtained after running $t$ LL iterations with a chosen first-order optimizer. Each optimizer step can be represented as the function $\psi_j : \mathbb{R}^n \times \mathbb{R}^m \to \mathbb{R}^n$, which gives the next iterate $\bs{x}_j = \psi_j(\bs{x}_{j-1}, \bs{y}), \, \forall j= \left\{ 1, \ldots, t \right\}$. The subscript on $\psi_j$ reflects the 
history dependence of the optimizer: the update at iteration $j$ 
may depend not only on $\boldsymbol{x}_{j-1}$ and $\boldsymbol{y}$, but 
also on accumulated quantities from previous iterations---such as 
momentum buffers and adaptive learning rate estimates---that together 
constitute the optimizer's internal state. The initial iterate may or may not depend on the UL variables, depending on the formulation; we include this dependence for completeness as $\boldsymbol{x}_0 = {\psi}_0(\boldsymbol{y})$.
Defining the Jacobians:
$$\mathbf{A}_j = \dfrac{\partial \psi_j}{\partial \bs{x}_{j-1}}=  \dfrac{\partial \bs{x}_j}{\partial \bs{x}_{j-1}}  \quad \text{and} \quad \mathbf{B}_j = \dfrac{\partial \psi_j}{\partial \bs{y}} = \dfrac{\partial \bs{x}_j}{\partial \bs{y}},$$which represent the partial derivatives of the optimizer step (at a given iteration) with respect to the previous iterate and UL variables, respectively. Note that the Jacobian $\mathbf{B}_j$ is non-zero whenever the UL variables enter the LL 
update---either through the LL objective $f$ (and hence its gradient), 
or through the optimizer's dynamics (e.g., step size, preconditioner, 
momentum coefficient). Assuming that $t$ is sufficiently large, chain rule on the whole trajectory gives a estimate of the \textit{best-response} Jacobian~\cite{shaban2019truncated}:

\begin{equation}
\begin{aligned}
    \frac{\partial \boldsymbol{x}_\star}{\partial \boldsymbol{y}} 
    \approx \frac{\partial \boldsymbol{x}_t}{\partial \boldsymbol{y}} 
    &= \sum_{j=0}^{t} \left( \prod_{k=j+1}^{t} \mathbf{A}_k \right) 
    \mathbf{B}_j \\
    &= \mathbf{B}_t + \mathbf{A}_t \mathbf{B}_{t-1} + 
    \mathbf{A}_t \mathbf{A}_{t-1} \mathbf{B}_{t-2} \\
    &\quad + \cdots + \underbrace{\mathbf{A}_t \mathbf{A}_{t-1} 
    \cdots \mathbf{A}_1}_{t \text{ terms}} \mathbf{B}_0
\end{aligned}
\label{eqn:adgrad}
\end{equation}
This sum denotes the sensitivity of a given iterate to $\bs{y}$ through each optimizer step; note that since ${\psi}_j$ uses the gradient of the LL objective, the Jacobians $\mathbf{A}_j$ and $\mathbf{B}_j$ contain 
second-order derivatives of $f$. As $t$ increases, this estimate converges to the true IFT hypergradient under mild conditions; even at small $t$, it remains informative 
because it captures the optimization trajectory rather than only the final iterate~\cite{lorraine2020optimizing, shaban2019truncated}.

For large $n$ and $m$, forming the Jacobian in Eqn.~\eqref{eqn:adgrad} 
explicitly is prohibitive. However, the hypergradient itself only requires the product 
of this Jacobian with vectors (Eqn.~\eqref{eqn:hpgrad}), which 
AD computes without ever materializing the Jacobian. 
This vector-Jacobian product can be obtained via either forward- or 
reverse-mode AD. Since the UL objective is a scalar function, forward-mode requires $m$ passes (one per hyperparameter) to construct the full hypergradient. Reverse-mode, by contrast, computes the hypergradient in a single backward pass and is therefore more efficient\footnote{Randomized forward mode estimators exist that can be both compute- and memory-efficient~\cite{shen2024memory}.}. 

To achieve this, the entire computation---from the UL variables 
through the LL optimization (including the optimizer dynamics) to the 
evaluation of the UL objective---must be expressed within an AD 
framework. Fig.~\ref{fig:ad_unroll} (top) shows a simplified 
computational graph, with each operation (optimizer steps, UL 
objective evaluation) represented by an orange node. Arrows denote 
the flow of data and show the dependencies; in particular, each 
optimizer step takes as input not only the previous iterate but also 
the UL variables $\boldsymbol{y}$.

In reverse-mode, the gradient signal is \textit{accumulated} by 
traversing the graph backwards, starting from the final output with a 
seed of $1.0$ (Fig.~\ref{fig:ad_unroll}, bottom). At each node, the local partial derivatives are computed 
and used to propagate the gradient to its inputs. A detailed treatment 
of this process in the TO context is given in~\cite{Sanu2026}; here 
we summarize the result. The entire reverse-mode computation can be 
expressed by two recursion rules, illustrated in Fig.~\ref{fig:ad_unroll}. Initializing 
$\bs{e}_t = \partial F / \partial \boldsymbol{x}_t$ and 
$\boldsymbol{c}_t = \partial F / \partial \boldsymbol{y}$---the latter 
being the direct partial term already encountered in 
Eqn.~\eqref{eqn:hpgrad}---the recursion proceeds from $j = t$ down to 
$1$:
\begin{equation}
    \boldsymbol{e}_{j-1} = \boldsymbol{e}_j \mathbf{A}_j, \qquad 
    \boldsymbol{c}_{j-1} = \boldsymbol{c}_j + \boldsymbol{e}_j \mathbf{B}_j,
    \label{eqn:revad}
\end{equation}
with the hypergradient given by $\boldsymbol{c}_0$ at the end of the 
backward pass. Although reverse mode is independent of the number of 
hyperparameters in computational cost, it requires storing the full 
LL trajectory of length $t$, incurring a memory cost proportional to 
$t \cdot n$~\cite{shaban2019truncated}.

\section{Methodology and implementation}

The UL objective $F(\boldsymbol{y}^{(k)}, \boldsymbol{x}^{(k)}_t)$ 
(where the superscript denotes the UL iteration and the subscript the 
LL iteration) depends not only on the current UL variables 
$\boldsymbol{y}^{(k)}$ but also on the final LL iterate 
$\boldsymbol{x}^{(k)}_t$. Because the LL is not run till convergence 
(finite $t$), this iterate is not necessarily the LL optimum but the endpoint of 
a trajectory, which depends on (1) the LL initialization 
$\boldsymbol{x}^{(k)}_0$, (2) the LL optimizer, (3) the horizon $t$, 
and (4) the UL variables $\boldsymbol{y}^{(k)}$. The UL variables 
affect the LL in two ways: some reshape the LL objective $f$ directly, while 
others (such as the learning rate or momentum) affect only how the 
optimizer traverses the landscape. We introduce four modifications targeting these components, to reduce 
computational cost and improve the stability of the bilevel 
optimization. We motivate and describe each below.

\begin{algorithm*}[t]
\caption{Both levels use Adam, with $\text{Adam}(\cdot)$ denoting 
    a single step; the LL optimizer's hyperparameters (e.g., learning rate) are contained in $\boldsymbol{y}_{\text{dyn}}$ and are themselves 
    tuned. The hyperparameters are partitioned as $\boldsymbol{y} = 
    (\boldsymbol{y}_{\text{loss}}, \boldsymbol{y}_{\text{dyn}})$, affecting the 
    LL loss landscape and optimizer dynamics respectively. The operators 
    $\text{sg}(\cdot)$ and $\text{clip}(\cdot, d_c)$ denote stop-gradient and 
    gradient clipping. UL stagnation is defined as both the EMA-smoothed 
    relative change in $F$ and the EMA-smoothed relative change in 
    $\boldsymbol{y}$ falling below user-specified thresholds (see 
    Appendix~\ref{app:hyperto_details} for values). LL optimizer state is 
    reinitialized at each UL iteration.}
\label{alg:hyperto}
\begin{algorithmic}[1]
\Require LL objective $f$, UL objective $F$; initial design 
    $\boldsymbol{x}^{(0)}_0$; initial hyperparameters $\boldsymbol{y}^{(0)}$ ; budget $\mathcal{B}_{\max}$; gradient 
    clipping threshold $d_c$; UL Adam learning rate $\eta^{\text{U}}$.
\Ensure $(\boldsymbol{y}^\star, \boldsymbol{x}_\star)$
\State $\mathcal{B} \gets 0$, \; $t \gets t_{\text{start}}$, \; $k \gets 0$, \; 
       $\boldsymbol{x}^{(-1)}_t \gets \boldsymbol{x}^{(0)}_0$
\While{$\mathcal{B} < \mathcal{B}_{\max}$}
    \State $\boldsymbol{x}^{(k)}_0 \gets \boldsymbol{x}^{(k-1)}_t$ 
        \Comment{Warmstart LL from previous UL iterate}
    \For{$j = 0, \dots, t-1$} \Comment{Run LL optimization for $t$ steps}
        \State $\bar{\boldsymbol{d}}^{(k)}_j \gets \text{clip}\!\left(
            \nabla_{\boldsymbol{x}} f\bigl(\text{sg}(\boldsymbol{x}^{(k)}_j); 
            \boldsymbol{y}_{\text{loss}}^{(k)}\bigr), d_c\right)$ 
            \Comment{Stop-gradient for pseudo first-order approx.}
        \State $\boldsymbol{x}^{(k)}_{j+1} \gets \text{Adam}\!\left(
            \boldsymbol{x}^{(k)}_j, \boldsymbol{d}^{(k)}_j, 
            \boldsymbol{y}_{\text{dyn}}^{(k)}\right)$
    \EndFor
    \State $F^{(k)} \gets F(\boldsymbol{y}^{(k)}, \boldsymbol{x}^{(k)}_t)$ 
        \Comment{Evaluate UL objective using final LL iterate}
    \State $\bar{\boldsymbol{d}}_{\boldsymbol{y}}^{(k)} \gets \text{clip}\!\left(
        \frac{d F^{(k)}}{d \bs{y}}, d_c\right)$
        \Comment{Clipped hypergradient through unrolled LL steps}
    \State $\boldsymbol{y}^{(k+1)} \gets 
        \mathcal{P}_{[\boldsymbol{y}_{\min}, \boldsymbol{y}_{\max}]}\!\left(
        \text{Adam}(\boldsymbol{y}^{(k)}, \boldsymbol{d}_{\boldsymbol{y}}^{(k)}, \eta^{\text{U}})
        \right)$
        \Comment{UL update with box projection}
    \If{UL stagnates} $t \gets \min(t + 1, t_{\max})$
    \EndIf 
    \State $\mathcal{B} \gets \mathcal{B} + (4t + 2)$, \; $k \gets k + 1$
\EndWhile
\State \Return $
    (\boldsymbol{y}^{(k)}, \boldsymbol{x}^{(k)}_t)$ 
    \Comment{Final iterates at termination}
\end{algorithmic}
\end{algorithm*}

\paragraph{First-order hypergradient}
From Eqn.~\eqref{eqn:adgrad}, and the definitions of $\mathbf{A}_t$ and $\mathbf{B}_t$, computing 
the hypergradient in general involves second-order terms of the LL objective 
$f$. To see this, consider gradient descent as the LL optimizer, giving the update rule $\bs{x}_{t} = \psi \left( \bs{x}_{t-1} \right) = \bs{x}_{t-1} - \eta  \nabla_{\bs{x}} f \left({\bs{x}_{t-1}} \right) $, 
for which:
\begin{align}
    \mathbf{A}_t &= \dfrac{\partial {\psi}_t}{\partial \bs{x}_{t-1}} 
         = \mathbf{I} - \eta \mathbf{H}_f({\bs{x}_{t-1}}), \\
    \mathbf{B}_t &= \frac{\partial {\psi}_t}{\partial \bs{y}} 
         = -\eta \dfrac{\partial^2 f}{\partial \bs{y}\, \partial 
           \bs{x}_{t-1}},
\end{align}
where $\eta$ is the learning rate or step size, $\mathbf{H}_f$ is the Hessian of $f$ evaluated at $\bs{x}_{t-1}$ and $\mathbf{I}$ is the 
identity matrix. To reduce computational cost, we set $\mathbf{H}_f = \bs{0}$, 
so that $\mathbf{A}_t \approx \mathbf{I}$ (In Fig.~\ref{fig:ad_unroll}, this corresponds to stopping gradient flow along the blue route) ~\cite{luketina2016scalable}. Notably, we keep the mixed partial term in $\mathbf{B}_t$, thereby preserving the dependence of the LL gradient on $\boldsymbol{y}$ and 
keeping the hypergradient estimate meaningful. This yields a pseudo first-order hypergradient, which is used throughout all 
experiments. While the above derivation uses gradient descent for 
clarity, the same approximation extends to other optimizers, 
including adaptive ones such as Adam~\cite{kingma2014adam}, which 
we use at both levels in all experiments.

\paragraph{Annealing the number of LL iterations}

Since the hypergradient is computed via AD, the number of LL 
iterations $t$ is a free algorithmic choice. Larger $t$ enforces LL 
optimality more strictly and yields a more accurate hypergradient, 
but at greater computational cost. A single UL update requires approximately $4t + 2$ evaluations of the 
LL objective $f$, which dominates the cost in TO\footnote{The $t$ LL updates contribute $2tf$ to the cost, since each step 
involves one objective evaluation and one adjoint-gradient computation 
of comparable cost. Computing the UL objective $F$ adds another $1f$, 
giving $2tf + 1f$ for the forward pass. The backward pass differentiates 
this entire computation: $1f$ for $F$ and $2tf$ for the $t$ LL updates, 
contributing $2tf + 1f$ more. Summing the two passes gives the total 
of $4t + 2$ function-equivalent evaluations.}. To keep 
this manageable, $t$ should remain small for most of the optimization.

Although $t$ is discrete and could in principle be relaxed to a 
continuous UL variable~\cite{jang2016categorical, zhang2022advancing}, 
we do not pursue this here. Instead, we adopt a simple \emph{annealing} 
strategy: the bilevel optimization is initialized with $t = t_{\text{start}}$ (typically set to $1$ or $2$), 
updating $\boldsymbol{y}$ after a single LL step. This gives a 
computationally cheap hypergradient estimate and supports the 
framework's \textit{nearly} simultaneous-optimization spirit, in which hyperparameter 
and design updates progress concurrently. When the UL fails to make meaningful progress---measured by the 
exponentially weighted moving averages of the relative changes in 
both the UL objective $F$ and the variables $\boldsymbol{y}$ falling 
below a threshold---$t$ is incremented by one, up to a prescribed 
maximum $t_{\max} = 10$.

Starting from $t = t_{\text{start}}$ can be seen as relaxing the LL optimality 
condition in the early iterations, smoothing the UL landscape and 
promoting broader exploration~\cite{metz2019understanding}, akin in 
spirit to continuation schemes in TO. However, operating at low $t$ 
introduces \textit{short-horizon bias}~\cite{metz2019understanding}: the UL 
objective is evaluated before LL convergence is reached, which can 
shift the apparent UL optimum and yield suboptimal LL solutions. 
Progressively increasing $t$ as the UL matures restores LL accuracy 
and mitigates this bias, as demonstrated empirically in 
Section~\ref{sec:smd}.

\paragraph{Warmstarting the LL initialization}
In the current bilevel formulation, the LL initialization is 
independent of the UL variables and must be chosen by the 
user\footnote{In some settings, such as meta-learning, the UL 
variables directly determine the LL initialization~\cite{finn2017maml}; 
we do not pursue that coupling here.}. The simplest choice is to 
initialize every LL solve from the same fixed starting point 
(\textit{coldstarting}). We instead use \textit{warmstarting}, in 
which the final LL iterate of the previous UL iteration becomes the 
initial point for the next LL solve~\cite{vicol2022implicit}:
\begin{align}
    \boldsymbol{x}^{(k)}_0 &= \boldsymbol{x}^{(0)}_0 \qquad \text{(coldstarting)}, 
    \label{eq:coldstart} \\
    \boldsymbol{x}^{(k+1)}_0 &= \boldsymbol{x}^{(k)}_t \qquad     \text{(warmstarting)}. 
    \label{eq:warmstart}
\end{align}

Both strategies apply to any bilevel solver, including ours; to 
isolate their effect, we hold everything else fixed---the LL horizon 
$t$ and the optimizer---so that the only difference is how each LL 
solve is initialized. The distinction is clearest by writing out the dependencies of the 
UL objective at iteration $k+1$:
\begin{equation}
    F^{(k+1)} = F\big(\boldsymbol{y}^{(k+1)}, \boldsymbol{x}^{(k+1)}_t\big),
\end{equation}
The two strategies differ only in how the initialization 
$\boldsymbol{x}^{(k+1)}_0$ is set at each iteration. Coldstarting fixes it at the same value across all 
iterations and hence function evaluations are completely independent. Warmstarting instead sets the current LL initialization to the 
previous endpoint (see Eqn.~\eqref{eq:warmstart}). Combined with the 
$t$-step LL optimization denoted as $\boldsymbol{x}^{(k+1)}_t = 
\boldsymbol{\Psi}(\boldsymbol{x}^{(k+1)}_0; \boldsymbol{y}^{(k+1)})$, 
this produces a recursion:
\begin{equation}
\begin{aligned}
    \boldsymbol{x}^{(k+1)}_t 
    &= \boldsymbol{\Psi}\big(\boldsymbol{x}^{(k+1)}_0; 
    \boldsymbol{y}^{(k+1)}\big) \\
    &= \boldsymbol{\Psi}\big(\boldsymbol{x}^{(k)}_t; 
    \boldsymbol{y}^{(k+1)}\big)  \\
    &= \boldsymbol{\Psi}\big(\boldsymbol{\Psi}(\boldsymbol{x}^{(k)}_0; 
    \boldsymbol{y}^{(k)}); \boldsymbol{y}^{(k+1)}\big)  \\
    &= \boldsymbol{\Psi}\big(\cdots \boldsymbol{\Psi}\big(
    \boldsymbol{x}^{(0)}_0; \boldsymbol{y}^{(0)}\big) \cdots; 
    \boldsymbol{y}^{(k+1)}\big).
\end{aligned}
\end{equation} 
The endpoint, and therefore $F^{(k+1)}$, now depends on the entire 
history $(\boldsymbol{y}^{(0)}, \ldots, \boldsymbol{y}^{(k+1)})$. The bilevel 
optimization becomes a dynamical system rather than a static 
optimization: the UL variables act as control inputs steering the 
joint state $(\boldsymbol{x}, \boldsymbol{y})$ along a trajectory. Under coldstarting, the UL objective is a static function of 
$\boldsymbol{y}$, and optimizing it yields a single point 
$\boldsymbol{y}^\star$. Under warmstarting, the objective is 
path-dependent, and the optimization instead traces out an ordered 
sequence $\{\boldsymbol{y}^{(k+1)}\}$, each element tuned for the 
warmstarted state at its iteration rather than for the cold 
initialization. We refer to this sequence as a \emph{learned hyperparameter 
schedule}.

Warmstarting offers two practical benefits. First, it enhances the 
smoothness of the map $\boldsymbol{y} \mapsto \boldsymbol{x}_t$: since 
the LL solver is initialized close to the previous solution, small 
changes in $\boldsymbol{y}$ produce small changes in the converged LL 
state. This smoothness controls the variance of the hypergradient and 
keeps it informative across UL iterations.  Second, it makes our annealing strategy viable. Using a small $t$ to 
estimate the hypergradient is only useful if 
meaningful progress toward the LL optimum can be made. Warmstarting ensures this 
since the LL 
optimality gap shrinks progressively across UL iterations. Without 
warmstarting, few LL step from a cold initialization would carry 
very little information about the LL optimum, and the annealing 
schedule's preference for small $t$ would fail. The progressive reduction of the LL 
optimality gap also improves the hypergradient estimate.

\paragraph{Gradient clipping and learning rate selection}

For the hypergradient to be well-behaved at a given UL iteration $k$, the mapping from the UL 
variables to the final LL state ($\boldsymbol{y}^{(k)} \mapsto \boldsymbol{x}_t^{(k)}$) 
must be smooth. This is achieved by keeping per-step updates small at 
both levels through two complementary mechanisms: gradient clipping 
and conservative learning rates.

Gradient clipping rescales the gradient vector when its norm exceeds 
a threshold $g_c$:
\begin{equation*}
\bar{\boldsymbol{d}} = 
\begin{cases}
    \dfrac{\boldsymbol{d}}{\|\boldsymbol{d}\|_2} \, d_c 
        & \text{if } \|\boldsymbol{d}\|_2 \geq d_c, \\[6pt]
    \boldsymbol{d} & \text{otherwise.}
\end{cases}
\end{equation*}
We apply it to the raw gradient at both levels ($\bs{d}$ can be either $\nabla f$ or $\frac{d F}{d \bs{y}}$) with $d_c$ set to $10$ times the initial gradient norm in 
all experiments. This serves primarily as a numerical safeguard 
against overflow, especially when many LL iterations are unrolled: 
the long chains of Jacobian products in Eqn.~\eqref{eqn:adgrad} may 
occasionally produce extreme gradient magnitudes (as we demonstrate empirically in Section~\ref{sec:smd}), and clipping 
prevents these from propagating into subsequent updates.

Although several first-order optimizers can be used with the framework, we use
Adam~\cite{kingma2014adam} at both levels for all experiments. Since we
differentiate through the LL, the LL Adam requires an additional numerical
safeguard. The update rule, for $t=1$, is
\begin{equation}
    \boldsymbol{x}_1 = \boldsymbol{x}_0 - \eta
    \frac{\bar{\boldsymbol{d}}}{\sqrt{\bar{\boldsymbol{d}}^2 + \tau_1} + \tau_2},
    \label{eqn:adam_t1}
\end{equation}
\noindent where all vector operations are element-wise and $\bar{\boldsymbol{g}}$
is the gradient after clipping. The constants $\tau_1 = \tau_2 = 10^{-8}$ guard
against division by zero. Here $\tau_1$ is placed inside the square root so that
the derivative of the square-root operation remains finite during AD through the
LL Adam updates. Since we do not differentiate through the UL update, this is
unnecessary there, and we set $\tau_1 = 0$ at the UL.

When each gradient component satisfies $|\bar{d}_i| \gg \sqrt{\tau_1} = 10^{-4}$,
the Adam denominator is approximately $|\bar{d}_i|$ and the update reduces to
$\boldsymbol{x}_0 - \eta\,\mathrm{sign}(\bar{\boldsymbol{d}})$
~\cite{abuduweili2025revisiting}: the step magnitude is controlled entirely by
$\eta$, independent of the gradient magnitude. This sign-descent behaviour has a
consequence for the hypergradient. For UL variables that affect the LL landscape
but have no direct term in the hypergradient (Eqn.~\ref{eqn:hpgrad}), $t=1$ is a
poor choice: because the $t=1$ update ignores gradient magnitude, the
hypergradient signal is nearly zero and dominated by noise. The suppression is
worst if Adam's moment estimates are reset at every UL iteration, which is what we
do. A simple remedy is to use $t=2$ instead.

Because $\eta$ largely determines the step magnitude, especially early in the
optimization, the LL learning rate requires care at initialization. A low initial
value provides stability but the framework may waste many UL iterations bringing
it up to an appropriate level; too high a value risks divergence. We therefore set
$\eta_0$ by a one-dimensional probe over a wide geometric grid on
$[\eta_{\min}, \eta_{\max}]$. Starting from the lower end of the bracket, we run
the LL optimizer for $t_{\text{start}}$ steps at each candidate and compare the
losses at the start and end of the run. A candidate is accepted if (i) it produces
a loss decrease, and (ii) that decrease stays below a target bound; the exact
criteria are given in Appendix~\ref{app:impl_lr_probe}. The probe uses exactly the
optimizer that is used later in the bilevel optimization, and each trial evaluates
the LL only, so no hypergradients are required. The framework subsequently adapts
$\eta$ via the hypergradient during the bilevel optimization. For the UL learning rate we use
$10^{-2}$ across all experiments unless stated otherwise, which controls the
convergence speed of the bilevel optimization.

\begin{figure*}[t]
	\centering
	\includegraphics[width=1.0\linewidth]{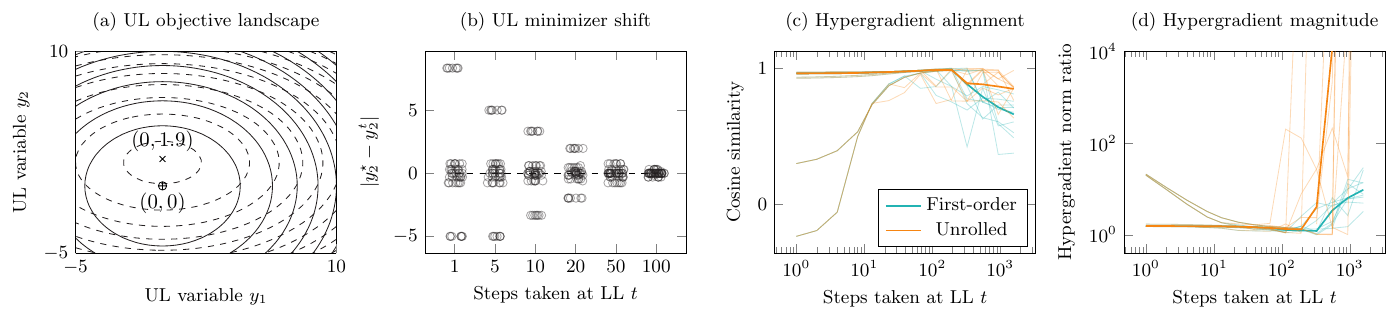}
	\caption{Illustration of challenges in gradient-based bilevel 
		optimization using the SMD-1 benchmark with two upper-level (UL) and 
		two lower-level (LL) variables. (a) True UL objective landscape 
		(solid contours), obtained using the analytical LL optimum at each 
		UL point, overlaid with an approximate landscape (dashed) obtained 
		from a single Adam step at the LL from a random initialization. The 
		true optimum at $(0,0)$ and the shifted apparent optimum are marked. 
		(b) Short-horizon bias: shift in the UL minimum from the true 
		location $y_2^\star$ to the apparent minimum $y_2^t$ when the LL is 
		run for $t$ steps, shown for several LL initializations. 
		(c) Cosine similarity between the estimated hypergradients (pseudo 
		first-order approximation and vanilla unrolling) and the true 
		hypergradient; dashed lines show individual runs and solid lines show 
		medians. (d) Ratio of the $L^2$ norms of the estimated and true 
		hypergradients, illustrating gradient norm explosion at large LL 
		horizons.}
	\label{fig:smd_part1}
\end{figure*}

The overall methodology is summarized in 
Algorithm~\ref{alg:hyperto}. Each UL iteration $k$ proceeds in three 
steps. First, the LL is warmstarted from the final iterate of the 
previous UL iteration ($\boldsymbol{x}^{(k-1)}_t$); this preserves 
proximity to the LL optimum across UL updates and reduces the number 
of LL steps required for convergence. Second, the LL is unrolled for 
$t$ steps of Adam, with the pseudo first-order approximation enforced 
via stop-gradient to keep the hypergradient computation tractable. 
Third, the hypergradient of the UL objective with respect to 
$\boldsymbol{y}$ is computed by reverse-mode AD through the unrolled 
trajectory and used to update $\boldsymbol{y}$, with gradient clipping 
and box-constraint projection for numerical safety. The LL horizon $t$ controls the cost of each UL iteration. We increase it adaptively when the UL optimization stagnates, 
up to a maximum $t_{\max}$. The total cost is tracked in cumulative 
LL iterations $\mathcal{B}_{\max}$, which provides a budget measure 
independent of the $t$-schedule and therefore enables fair comparison 
across different bilevel runs. More details on the numerical 
implementation are given in Appendix~\ref{app:hyperto_details}.

\begin{figure*}[t]
	\centering
	\includegraphics[width=1\linewidth]{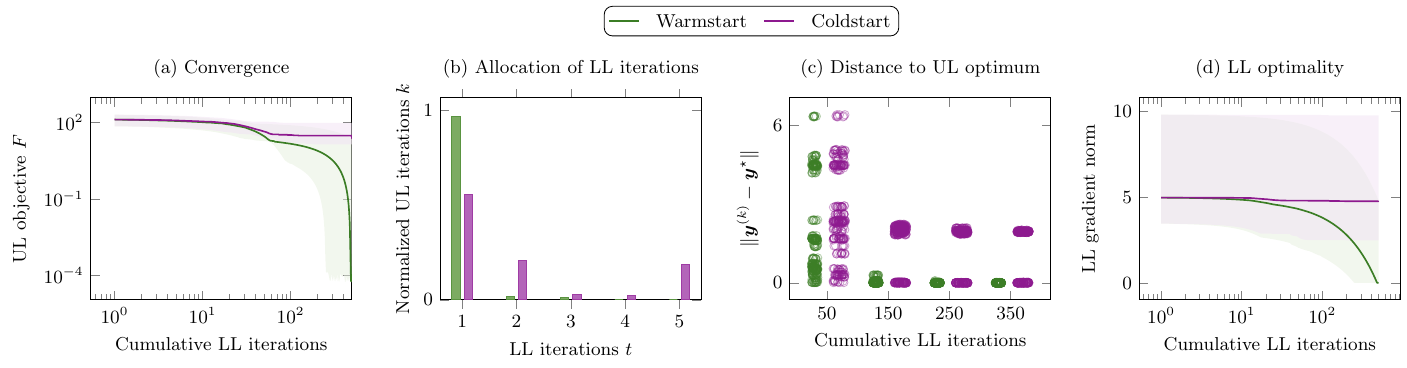}
	\caption{Effect of warmstarting on the bilevel optimization of 
		SMD-1. All four panels compare warmstarting and coldstarting under 
		otherwise identical settings (Adam at both levels, gradient clipping, 
		annealing schedule). Results are aggregated over $72$ runs combining 
		$9$ UL and $8$ LL initializations; solid lines show medians and 
		shaded bands show one standard deviation. (a) Convergence history 
		of the UL objective; the horizontal axis is the cumulative number 
		of LL steps. (b) Fraction of UL iterations spent at each value of 
		the LL horizon $t$, illustrating how the annealing schedule 
		allocates the budget. For each scheme, the bar heights 
		sum to one. (c) Euclidean distance between the UL 
		iterate $\boldsymbol{y}^{(k)}$ after $k$ UL iterations and the true UL optimum 
		$\boldsymbol{y}^\star = (0, 0)$. (d) LL optimality measured by the 
		norm of the LL objective gradient $\|\nabla_{\boldsymbol{x}} f\|$ 
		at the final LL iterate.}
	\label{fig:smd_part2}
\end{figure*}

\section{Verification on a benchmark bilevel problem} \label{sec:smd}

To illustrate the challenges of hypergradient-based bilevel 
optimization and demonstrate how the proposed framework addresses them, we first 
consider a bilevel optimization test problem that has an analytical solution. Specifically, we borrow a problem from the SMD test suite~\cite{Sinha2014}, which is a collection 
of benchmark problems designed to exhibit characteristic difficulties 
encountered in bilevel optimization. Unlike practical applications, these benchmark problems possess analytically tractable LL optima, allowing the exact UL objective landscape and corresponding hypergradients to be computed. This makes it possible to directly compare the true UL landscape against the approximate landscape obtained when the LL optimization is solved only approximately using a finite number of optimization iterations.

The bilevel optimization problem SMD-1, which is the simplest problem in the SMD suite, is given by

\begin{equation} \label{eq:smd1}
\begin{aligned} 
\min_{\boldsymbol{y}} \;  F  &= \|\boldsymbol{y}_{1}\|^{2} + \|\bs{x}_{1\star}\|^{2} + \|\boldsymbol{y}_{2}\|^{2}\\
& \qquad + \|\boldsymbol{y}_{2} - \tan(\bs{x}_{2\star})\|^{2} \\[0.5em] 
\text{s.t.} &\;  \bs{x}_\star = \argmin_{\bs{x}} \; f = \|\boldsymbol{y}_{1}\|^{2} + \|\bs{x}_{1}\|^{2} \\
& \qquad \hspace{65pt} + \|\boldsymbol{y}_{2} - \tan(\bs{x}_{2})\|^{2} \\[0.5em] 
&\; \begin{array}{r@{\;}c@{\;}l@{\qquad}l} -5 &\le& (\boldsymbol{y}_{1})_i \le 10,  & i = 1,\dots,p,\\ -5 &\le& (\boldsymbol{y}_{2})_i \le 10,  & i = 1,\dots,r,\\ -5 &\le& (\bs{x}_{1})_i \le 10,  & i = 1,\dots,q,\\ -\pi/2 &<& (\bs{x}_{2})_i < \pi/2,  & i = 1,\dots,r, \end{array} 
\end{aligned}
\end{equation}

\noindent where the UL and LL variables are each partitioned into two groups, 
$\boldsymbol{y} = (\boldsymbol{y}_1, \boldsymbol{y}_2)$ and 
$\boldsymbol{x} = (\boldsymbol{x}_1, \boldsymbol{x}_2)$, with the 
second group of each level coupling the two problems and the first 
group adding within-level complexity. Problem~\eqref{eq:smd1} has a convex UL objective and a unimodal LL landscape for 
each UL point\footnote{Although the original work reported the LL 
problem to also be convex, our verification shows this holds only when 
$\boldsymbol{y}_2 \in [-\sqrt{3}, +\sqrt{3}]$, as confirmed by 
checking positive semi-definiteness of the LL Hessian over this range.}. 
Despite its simplicity, this problem already exhibits several challenges 
characteristic of gradient-based bilevel optimization, suggesting that 
similar difficulties will be more pronounced in TO problems involving 
high-dimensional, nonlinear physical simulations.

Fig.~\hyperref[fig:smd_part1]{\ref*{fig:smd_part1}a} shows isocontours 
of the true UL objective landscape (solid lines), where the UL 
variables are denoted $y_1$ and $y_2$. Since the LL optimum is known 
analytically, the true UL landscape is a function of $\boldsymbol{y}$ 
alone. We sample $F$ on a 
$40 \times 40$ uniform grid over the range to produce the isocontours shown. The resulting landscape 
is convex, with the global optimum at $\boldsymbol{y}^\star = 
(0, 0)$. Overlaid on the same figure is an approximate UL landscape (dashed 
isolines), obtained by replacing the exact LL solution 
$\boldsymbol{x}_\star$ with the finite-step approximation 
$\boldsymbol{x}_t$. For each grid point, we initialize the LL at a 
random point (held fixed across the grid) and take a single Adam step 
($t = 1$ with learning rate $10^{-2}$) to compute $\boldsymbol{x}_t$. As shown in the figure, a shift in the apparent UL minimum is visible along $y_2$, which is the only UL variable coupled to the LL optimum through the term $\|\boldsymbol{y}_2 - \tan(\bs{x}_{2\star})\|^2$. The other component ($y_1$) enters the UL objective independently of the LL solution. This phenomenon, whereby the UL optimum is shifted due to the lack of convergence in the LL problem, is known as \emph{short-horizon bias}~\cite{wu2018understanding}. We therefore see that the approximate UL objective therefore depends not only on the UL variables, but also on the LL optimization trajectory, its initialization, and the number of iterations performed---none of which affect the true UL landscape.

Fig.~\hyperref[fig:smd_part1]{\ref*{fig:smd_part1}b} shows how this 
bias varies as more LL steps are taken. To produce the plot, we 
consider $50$ random LL initializations. For each, we evaluate the 
UL objective densely along $y_2$ ($100$ grid points) 
keeping $y_1$ fixed at $0$ and identify the apparent minimum 
$y_2^t$. This procedure is repeated for varying $t$. 
The bias ($|y_2^\star - y_2^t|$) decreases with $t$ and vanishes at sufficiently large $t$, 
as $\boldsymbol{x}_t$ converges to $\boldsymbol{x}_\star$ and the 
approximate landscape coincides with the true one.

Reducing the bias, however, comes at a dual cost: increased 
computational expense and a potential deterioration in hypergradient 
quality. The latter is illustrated in 
Fig.~\hyperref[fig:smd_part1]{\ref*{fig:smd_part1}c}, which compares 
the cosine similarity between the true hypergradient---computed 
analytically via Eqn.~\eqref{eqn:uncongrad}---and the estimates 
obtained from Eqn.~\eqref{eqn:adgrad}, both with and without the 
pseudo first-order approximation. For each value of $t$, we sample 
$225$ uniformly distributed UL points and compute the hypergradient 
from $100$ random LL initializations per UL point, since the 
estimates depend on the LL initialization. Although both estimators 
initially agree well with the true hypergradient, accuracy degrades 
as $t$ grows~\cite{metz2021gradients, metz2019understanding}. This 
degradation arises because AD-based estimators involve products of 
Jacobian matrices of increasing length, which become numerically 
unstable as $t$ grows.

Fig.~\hyperref[fig:smd_part1]{\ref*{fig:smd_part1}d} confirms that 
this numerical instability is in fact responsible for the 
degradation. For the same points as in the previous plot, we evaluate 
the ratio between the $L^2$ norm of the estimated hypergradients and 
the true hypergradient. The ratio becomes extremely large, exceeding 
the true gradient magnitude by several orders of magnitude. The 
pseudo first-order estimator is more stable because it eliminates the Hessian terms whose 
accumulation through the unrolled trajectory drives part of the amplification (see Eqn.~\eqref{eqn:adgrad}). 
This behavior motivates the use of gradient clipping as a practical 
safeguard.

The issues highlighted above can be addressed by the proposed 
framework---pseudo first-order hypergradients, gradient clipping, 
annealing, and warmstarting. Of particular importance is the use of warmstarting rather than 
coldstarting at each UL iteration. Recall that, by coldstarting we mean that the 
LL initialization is fixed at the same user-chosen value across all 
UL iterations. Fig.~\ref{fig:smd_part2} compares the two strategies on SMD-1; aside 
from the choice of warmstarting versus coldstarting, all other 
algorithmic details are kept identical (Adam at both levels, gradient 
clipping, and the annealing schedule described earlier). The experiment uses $72$ runs corresponding to all combinations of 
$9$ UL initializations and $8$ LL initializations sampled on a grid 
spanning the UL and LL ranges. We start the bilevel optimization 
from each of these combinations. As is evident from Fig.~\hyperref[fig:smd_part2]{\ref*{fig:smd_part2}a}, warmstarting converges significantly faster and reaches 
lower UL objective values than coldstarting with medians shown as 
thick lines, and the shaded bands indicating one standard deviation.

Fig.~\hyperref[fig:smd_part2]{\ref*{fig:smd_part2}b} shows how the 
annealing schedule allocates the total budget of $1500$ LL iterations 
(here with $t_{\max} = 5$). The horizontal axis is $t$ and the 
vertical axis reports the fraction of UL iterations spent at each $t$ 
value. The majority of UL iterations operate at $t = 1$, confirming 
that even the cheapest hypergradient estimate carries useful 
directional information. Fig.~\hyperref[fig:smd_part2]{\ref*{fig:smd_part2}c} shows that 
warmstarting consistently drives the solution closer to the true UL 
optimum, while coldstarting frequently stagnates in suboptimal 
regions. This is a consequence of coldstarting encountering more 
non-convexities at each UL iteration, as illustrated in 
Appendix~\ref{app:smd_details}. Finally, Fig.~\hyperref[fig:smd_part2]{\ref*{fig:smd_part2}d} shows 
that warmstarting enables simultaneous progress toward LL optimality, 
whereas coldstarting fails to reduce the LL gradient norm reliably.

\section{Bilevel compliance minimization}
\label{sec:compliance}

We now demonstrate the proposed framework in the context of compliance minimization, a standard benchmark problem in density-based TO~\cite{sigmund200199, ferrari2020new, andreassen2011efficient}. Although the formulation contains a constraint, feasibility is enforced implicitly through projection operations, keeping the LL optimization effectively unconstrained. The TO problem, which forms the LL, is formulated as:

\begin{equation}
\begin{aligned}
\bs{x}_{\star} &= \argmin_{\bs{x} \in \mathbb{R}^M} \quad 
 f = \bs{u}^{\intercal}\bs{f}, \\ 
& \text{s.t.} \quad 
 \mathbf{K}(\boldsymbol{\rho})\,\bs{u} = \bs{f}, \\
& \tilde{\boldsymbol{\rho}} = \Phi(\bs{x}) \in [0,1]^N, \quad \text{with} \quad \sum_{i=1}^{N} v_i \tilde{\rho}_i = V_0,  \\
& \boldsymbol{\rho} = \mathcal{F}(\tilde{\boldsymbol{\rho}}),
\end{aligned}
\label{eq:compliance_LL}
\end{equation}
where $\boldsymbol{u} \in \mathbb{R}^N$ is the static equilibrium displacement vector obtained via finite element analysis, 
$\boldsymbol{f}$ is the force vector, and $\mathbf{K}$ is the global 
stiffness matrix. The objective $f$ measures the structure's compliance. The stiffness matrix depends on the element-wise physical 
density field ${\boldsymbol{\rho}}$ via the solid isotropic 
material with penalization (SIMP) interpolation:
\begin{equation}
    E({\rho}_e) = E_{\min} + (E_0 - E_{\min}) \, {\rho}_e^p,
    \label{eq:simp}
\end{equation}
where $p \geq 1$ is the SIMP penalty parameter, $E_0 = 1.0$ is the 
solid material's Young's modulus, and $E_{\min} = 10^{-6}$ is the 
void stiffness.

The physical density field ${\boldsymbol{\rho}}$ is obtained 
from the optimization variables $\boldsymbol{x}$ through a chain of operations: starting with a parametric sigmoidal projection $\Phi$ followed by a series of filters $\mathcal{F}$. In the \textit{standard parameterization}, $\boldsymbol{x}$ is 
itself an element-wise density field with $M = N$ (the number of 
finite elements). In the \textit{neural topology optimization} 
parameterization~\cite{hoyer, Chandrasekhar2020}, $\boldsymbol{x}$ 
is the weights and biases of an untrained neural network whose 
output is the element-wise density field. In both cases, this field 
(not yet the physical density) serves as the input to the subsequent 
projection and filtering operations.
Such a reparameterization distorts the landscape---often making it 
non-convex---and substantially increases sensitivity to hyperparameter 
selection~\cite{Sanu2025}, providing a challenging test for the 
framework.

\begin{figure}[h]
	\centering
	\includegraphics[width=1\linewidth]{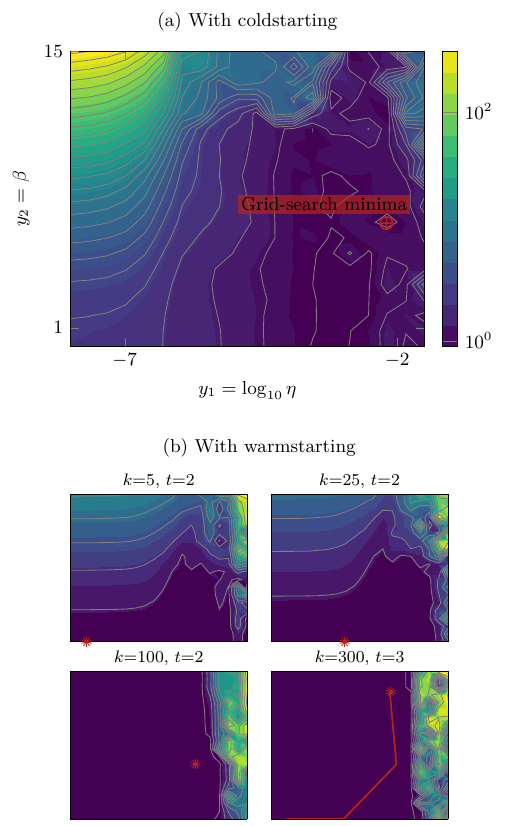}
	\caption{The UL landscape for neural topology optimization with a CNN architecture. (a) Landscape obtained by independently solving the lower-level problem from the same fixed initialization for each hyperparameter pair $(\eta,\beta)$, corresponding to the landscape observed by BO. (b) Effective landscape encountered by BOTH under warmstarted optimization, illustrating the path-dependent nature of hypergradient-based tuning; the red asterisk marks the current UL iterate and the trailing 
		line shows the trajectory followed.. Here, $k$ denotes the UL iteration index and $t$ denotes the number of LL optimization steps performed at that iteration.}
	\label{fig:comp2d_landscape}
\end{figure}

The values entering $\Phi$ (from either parameterization) are then mapped to 
$[0, 1]$ and the sigmoid is adaptively shifted via 
bisection so that the global volume constraint is satisfied at every 
iteration~\cite{hoyer}. This enables the use 
of unconstrained optimizers such as Adam at the LL. The projected densities $\tilde{\boldsymbol{\rho}}$ are then passed 
through a filtering stage $\mathcal{F}$, which consists of two 
operations. First, a density filter smoothens the design field to prevent 
checkerboard patterns and enforce a minimum length 
scale~\cite{bruns2001topology, bourdin2001filters}:
\begin{equation}
\label{eq:density_filter}
\begin{aligned}
\bar{\rho}_e &= \frac{\sum_{i \in \mathcal{N}_e} h_{ei}\, \tilde{\rho}_i}
                     {\sum_{i \in \mathcal{N}_e} h_{ei}}, \\
h_{ei} &= \max\bigl(0,\, r_{\min} - \mathrm{dist}(\Omega_i, \Omega_e)\bigr),
\end{aligned}
\end{equation}
where $\mathcal{N}_e = \{i \mid \mathrm{dist}(\Omega_i, \Omega_e) \leq 
r_{\min}\}$ is the neighborhood of element $\Omega_e$, 
$\mathrm{dist}(\cdot, \cdot)$ denotes the centre-to-centre distance, 
and $r_{\min}$ is the filter radius. Second, a Heaviside projection 
sharpens the filtered field $\bar{\rho}_e$ to remove intermediate 
densities~\cite{Xu2009}:
\begin{equation}
\label{eq:heaviside}
    \rho_e = \frac{\tanh(\beta \eta_h) + 
                   \tanh\!\bigl(\beta(\bar{\rho}_e - \eta_h)\bigr)}
                  {\tanh(\beta \eta_h) + 
                   \tanh\!\bigl(\beta(1 - \eta_h)\bigr)},
\end{equation}
where $\beta$ controls the sharpness of the projection and $\eta_h$ 
is the threshold, either determined via bisection to enforce volume 
conservation across the projection or fixed to $0.5$.

\begin{figure*}[h]
	\centering
	\includegraphics[width=1\linewidth]{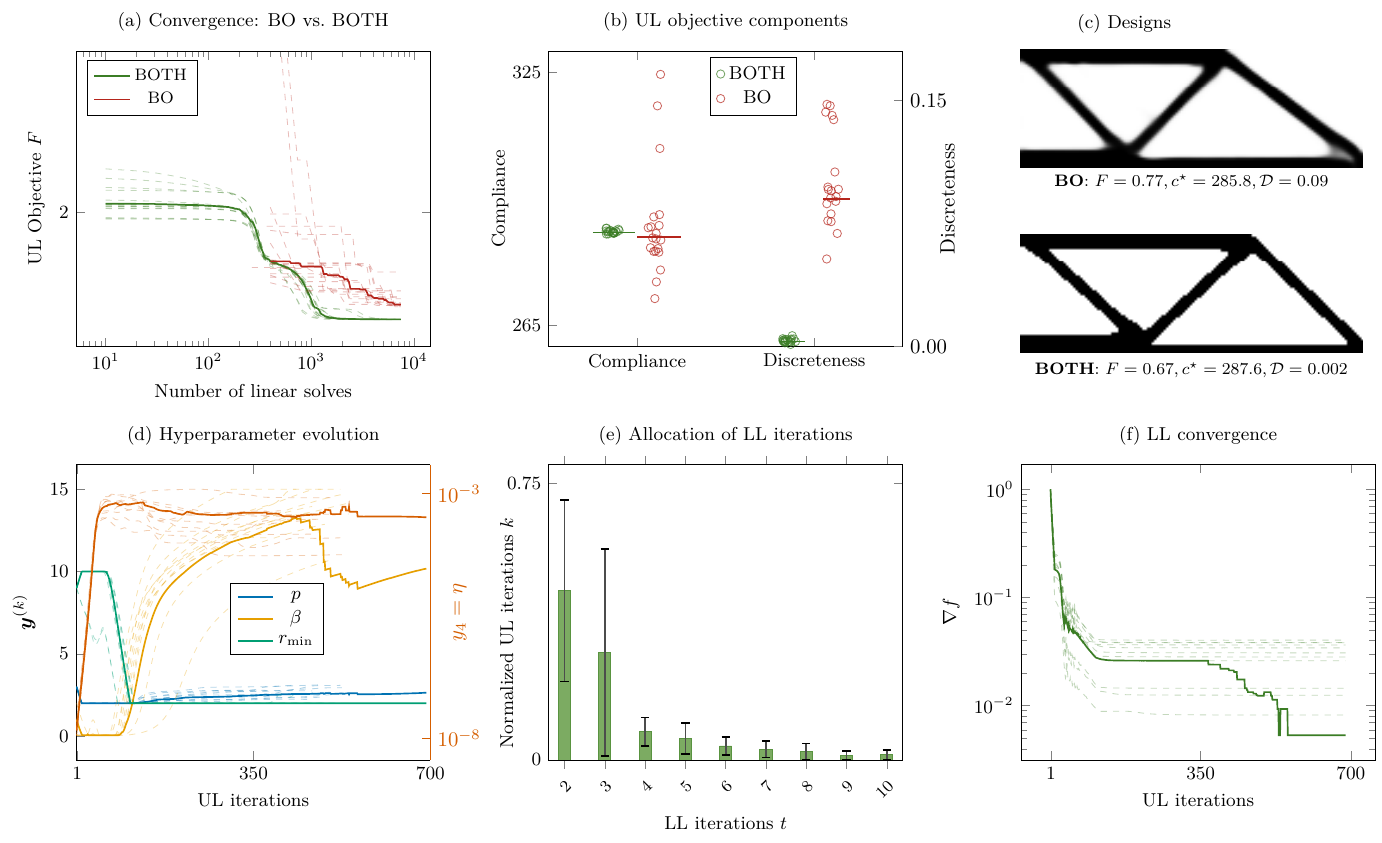}
	\caption{Comparison between Bayesian optimization (BO) and our method (BOTH) 
		for neural topology optimization with a CNN architecture. (a) 
		Convergence histories as a function of cumulative number of linear solves used by each method.
		Dashed lines show individual runs from different initializations; 
		the bold line shows the median. (b) Decomposition 
		of the final UL objective into its compliance and discreteness 
		components across runs, with medians marked by horizontal lines. 
		(c) Typical topologies obtained for the MBB beam together with their 
		UL objective values, where $c^\star$ denotes compliance (evaluated at $p=1$) and $\mathcal{D}$ measures discreteness. (d) 
		Evolution of the four tuned hyperparameters: learning rate $\eta$, 
		Heaviside sharpness $\beta$, SIMP penalty exponent $p$, and filter 
		radius $r_{\min}$; $\eta$ is shown on the right $y$-axis. (e) Mean 
		number of UL iterations (normalized against the total UL iterations) spent at each LL horizon $t$, averaged 
		across all runs, with one standard deviation shown as error bars; 
		$k$ denotes the UL iteration index. (f) LL objective value at the 
		final LL iterate $f(\boldsymbol{x}_t^{(k)})$ as a function of UL 
		iteration $k$.}
	\label{fig:comp4d}
\end{figure*}

Following~\cite{Ha2024}, we define the UL objective as a weighted combination of compliance and discreteness:
\begin{equation}
\begin{aligned}
    F(\boldsymbol{y}, \boldsymbol{x}_\star) 
    = \frac{c(\boldsymbol{x}_\star)}{c_{\text{ref}}} + k \, \frac{\mathcal{D}({\boldsymbol{\rho}}_\star)}{d_{\text{ref}}}, \; \text{with} \;
    \\
    \mathcal{D}({\boldsymbol{\rho}}_\star) = 
    \frac{\sum_{e=1}^N 4 \, {\rho}_{\star, e} \left(1 - {\rho}_{\star, e} \right) v_e}
         {\sum_{e=1}^N v_e},
\end{aligned}
\label{eq:compliance_UL}
\end{equation}
\noindent where $\mathcal{D} \in [0, 1]$ measures the discreteness of 
the design ($\mathcal{D} = 0$ for a fully discrete 
design)~\cite{sigmund2007morphology}, $v_e$ is the volume of element 
$e$, and $k$ is a user-chosen weight. The physical density 
${\boldsymbol{\rho}}_\star$, on which FEA is performed, is 
obtained from $\boldsymbol{x}_\star$ through the projection and 
filtering operations defined earlier.The compliance $c(\boldsymbol{x}_\star)$ is evaluated on the terminal LL design
with the penalization exponent set to $p = 1$; every other quantity entering the
evaluation is taken at its current LL value. 
A volume violation penalty is not required, since the volume 
constraint is enforced exactly at every LL iteration through the 
adaptive projection threshold in $\Phi$. We fix $k = 1$ and normalize both terms to near-equal magnitude at 
initialization using ${c_{\text{ref}}}$ and ${d_{\text{ref}}}$; an ablation 
study on $k$ and $\beta$ at the UL is given in 
Appendix~\ref{app:comp}.
 
\subsection{Results}
\label{subsec:compliance_bilevel}

We begin by selecting two UL variables: the LL optimizer's learning rate 
$\eta$ and the Heaviside projection sharpness $\beta$. This 
two-dimensional hyperparameter space allows us to visualize the UL 
landscape. All other hyperparameters such as the SIMP penalty and the filer radius are fixed. We present results for the over-parameterized convolutional neural network
architecture proposed by~\cite{hoyer}; results for the standard 
element-wise parameterization are provided in 
Appendix~\ref{app:comp}\footnote{The landscape for CNN is notably more 
challenging than the one obtained under the standard parameterization 
(Fig.~\hyperref[fig:app_comp2d_landscape]{\ref*{fig:app_comp2d_landscape}a}).}.

The 
UL landscape in Fig.~\hyperref[fig:comp2d_landscape]{\ref*{fig:comp2d_landscape}a} is 
obtained by sampling $\boldsymbol{y} = (\log_{10}\eta, \beta)$ on a 
$20 \times 20$ grid within the range $[-8.0, -1.5] 
\times [0.1, 15]$ and evaluating $F$ after $t = 150$ 
LL Adam steps from a fixed initialization at each 
grid point. Methods such as BO evaluate each hyperparameter 
configuration independently (from the same network initialization 
$\bs{x}^{(0)}$), and therefore optimize over precisely this static 
landscape.

BOTH, by contrast, uses warmstarting, so the 
landscape it effectively traverses evolves over the course of the 
optimization and depends on the trajectory taken. To visualize this, 
we run an actual bilevel optimization and take snapshots at several 
points along its trajectory. At each snapshot $k$, we hold the 
warmstarted state $\boldsymbol{x}^{(k)}_0$ fixed and reconstruct the 
landscape by sampling the same $20 \times 20$ grid, this time 
evaluating $F$ after $t$ LL steps (with $t$ set by the annealing 
schedule at that snapshot rather than the fixed $t = 150$ used for the 
static landscape). Fig.~\hyperref[fig:comp2d_landscape]{\ref*{fig:comp2d_landscape}b} 
shows the resulting sequence of landscapes. Unlike the non-convex 
static landscape in Fig.~\hyperref[fig:comp2d_landscape]{\ref*{fig:comp2d_landscape}a}, 
these warmstarted landscapes are noticeably smoother near the 
trajectory, consistent with the reduced non-convexity quantified in 
Fig.~\ref{fig:warm_ncr}.

We compare BOTH against BO, a strong baseline for low-dimensional 
hyperparameter optimization. We tune four hyperparameters: the LL 
learning rate $\eta$, the Heaviside sharpness $\beta$, the SIMP 
penalty exponent $p$, and the filter radius $r_{\min}$. BO proceeds 
by sampling an initial set of configurations uniformly at random, 
evaluating each by running TO to convergence (convergence criteria 
given in Appendix~\ref{app:comp}), and computing the UL objective. A 
Gaussian process surrogate is then fit to the observed 
configuration--performance pairs, and an acquisition function is 
optimized to propose the next configuration. Each evaluation updates 
the surrogate, and the procedure repeats until the budget is 
exhausted. Since the number of LL iterations per BO sample varies 
with convergence speed, cumulative LL iterations serves as the 
common basis for comparison (further details in 
Appendix~\ref{app:comp}).

The convergence histories for both methods are shown in 
Fig.~\hyperref[fig:comp4d]{\ref*{fig:comp4d}a}. The objective values $F$ are comparable since we use the same reference values for all runs. Since BO's initial 
sampling is stochastic, we repeat it with $4$ random seeds and five 
neural network initializations per seed, giving $24$ runs in total. 
For BOTH, $20$ runs are obtained from $5$ neural network 
initializations crossed with $4$ values of $\beta^{(0)}$; the UL 
optimizer is Adam with learning rate $10^{-1}$. Dashed lines show 
individual runs and thick curves show the median across runs. Unlike 
BOTH, BO curves do not begin near zero on the $x$-axis: each BO 
sample requires a full TO run to convergence (roughly $100$ LL 
iterations) before contributing a single data point, so the first BO 
observation appears only after this fixed startup cost.

BOTH produces designs with lower compliance and discreteness 
penalty, while BO produces a wider spread of final values 
(Fig.~\hyperref[fig:comp4d]{\ref*{fig:comp4d}b}), with a 
representative design shown in 
Fig.~\hyperref[fig:comp4d]{\ref*{fig:comp4d}c}. Even in this 
low-dimensional setting where BO is expected to be competitive, 
BOTH converges to a lower UL objective within the same 
computational budget.

The hyperparameter schedules $\boldsymbol{y}^{(k)}$ found by HyperTO 
are shown in Fig.~\hyperref[fig:comp4d]{\ref*{fig:comp4d}d} for a 
subset of runs, with the median plotted in thick lines. The learning 
rate schedule exhibits an initial increase to accelerate early 
progress, followed by a gradual decay to stabilize convergence. The 
learned $\beta$ remains small for most of the optimization, increasing 
only in the later stages; this allows the optimizer to first explore 
smoother intermediate density fields before enforcing discreteness 
near convergence. The evolution of $r_{\min}$ serves as a sanity 
check: for compliance minimization, the filter radius should be as 
small as possible, and indeed $r_{\min}$ converges to its lower bound 
(set to $r_{\min} = 2.0$ to prevent checkerboard patterns). Finally, 
runs from different initializations follow distinct schedules, 
indicating that BOTH adapts to each run's specific optimization 
trajectory rather than converging to a single universal schedule.

Fig.~\hyperref[fig:comp4d]{\ref*{fig:comp4d}e} shows the fraction of 
UL iterations spent at each annealing stage $t$, confirming that the 
budget is concentrated at low $t$ values. This supports the argument 
that even a small number of LL steps produces a sufficiently 
informative hypergradient estimate. 
Fig.~\hyperref[fig:comp4d]{\ref*{fig:comp4d}f} shows the evolution 
of the LL objective as a function of UL iterations $k$, recording the 
final LL value at each stage (i.e., $f(\boldsymbol{x}_t^{(k)})$). 
The LL objective decreases by approximately an order of magnitude 
before plateauing, consistent with the LL solve being warmstarted 
from a near-optimal point at each UL iteration.

\begin{figure}[]
    \centering
    \includegraphics[width=1\columnwidth]{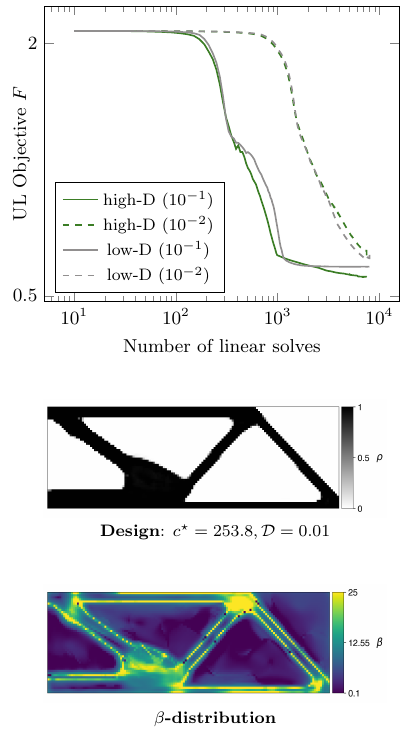}
\caption{Scaling to a high-dimensional UL via per-element projection 
sharpness. Each element is assigned its own $\beta_e$, giving 6915 UL 
hyperparameters (high-D, green), compared against the earlier 
single-$\beta$ case (low-D, gray), each shown for two UL Adam 
learning rates. The high-D and low-D cases converge comparably at 
similar per-iteration cost. Bottom: a representative final design and 
the corresponding learned $\beta_e$ field.}
\label{fig:compnd}
\end{figure}

We extend the previous example by assigning each element its own 
projection sharpness $\beta_e$ instead of a single global $\beta$, 
yielding a high-dimensional (high-D) UL with $6915$ variables. 
This scale is only tractable because the hypergradient cost is 
independent of the UL dimension---a regime inaccessible to 
derivative-free tuning. Fig.~\ref{fig:compnd} contrasts the high-D 
convergence (green) against the earlier low-D case (black), for two 
UL Adam learning rates. The high-D case converges comparably to the 
low-D case, at nearly the same per-iteration cost, demonstrating the 
framework's scaling. The learned $\beta_e$ field concentrates high 
values along the edges, consistent with intuition: sharper projection 
there counteracts the grayness introduced by the density filter.

\section{Constrained LL optimization} \label{sec:LL_constraints}
To showcase the use of hypergradients in the presence of LL non-trivial stress constraints, we first consider a canonical two-bar benchmark problem from~\cite{Verbart2016}. The problem consists of two bars connected at a shared node where an axial force is applied, while the left and right end nodes are fixed (see Fig.~\ref{fig:stress2d_qpm}). These loading and boundary conditions induce tensile stress in one bar and compressive stress in the other, making the example a simple yet instructive setting for studying stress-constrained optimization. The cross-sectional areas of the two bars ($\bs{a} = \left(a_1, a_2\right)$) are the design variables. We seek to minimize the total mass of the structure subject to upper bounds on the magnitude of the stresses. The formulation is given by:
\begin{figure*}[t]
    \centering
    \includegraphics[width=1\textwidth]{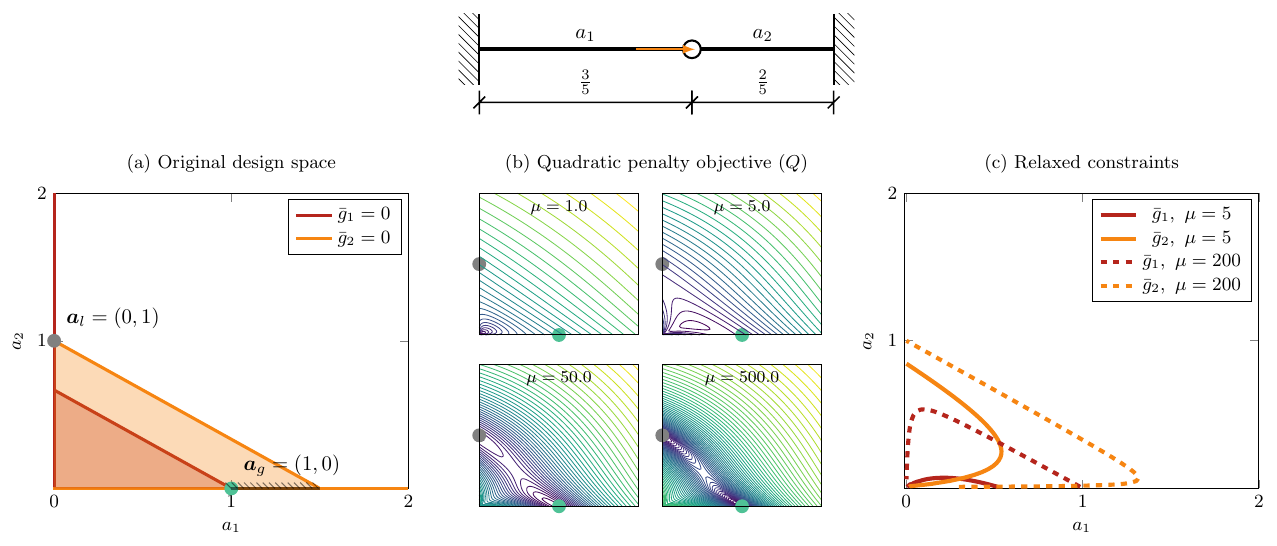}
\caption{Two-bar stress-constrained optimization benchmark problem, borrowed
from~\cite{Verbart2016}. (top) Schematic of the structure and boundary conditions for the two dimensional toy problem, with the areas of the bars $\bs{a} = (a_1, a_2)$ being the design variables. Load is applied axially at the point of contact. (a) The problem's design space, outlining the two vanishing stress constraints ($\bar{g}_1, \bar{g}_2$). Note that the constraint boundaries also include the coordinate axes. The main feasible region (in white) has an one-dimensional extension along the $a_1$ axis (hashed) allowing access to the singular global optimum at $(1,0)$. (b) The augmented objective of the quadratic penalty method, consisting of the mass and the sum of the squares of the constraints for different values of the penalty factor ($\mu$) is shown. (c) The equivalent relaxed constraints (using the optimal Lagrange multipliers at $(1,0)$) for two values of $\mu$.}
\label{fig:stress2d_qpm}
\end{figure*}

\begin{equation} \label{two_bar_problem}
	\begin{aligned}
		\bs{a}_{\star} = \argmin_{ \bs{a} \in \mathbb{R}^2} \quad &  f\left( a_1, a_2 \right) = 0.6a_1 + 0.8a_2, \\
		\text{such that} \quad & \bar{g}_i = \textstyle \left( \frac{a_i}{2} \right)g_i \leq 0, \\
		\quad & 0 \leq a_i \leq 2, \quad i \in \left\{ 1, 2 \right\}, \\
	\end{aligned}
\end{equation}
where the unscaled stress constraint is $g_i = \frac{\lvert \sigma_i \rvert}{\sigma_{\max}} - 1$, and $\sigma_{\max} = 1$ is the allowable stress (all details pertaining to this problem are detailed in Appendix~\ref{app:stress2d}). This scaling of the constraints with the design variables ensures that the constraint function is ``design-independent'', so that the function is defined over the whole design domain and not just where material exists (from $a_i > 0$ to $a_i\geq0$)~\cite{Rozvany2001}. This also helps the constraints vanish when the area is zero, and such constraints are thus known as \textit{vanishing constraints}. However, the optimization problem~\eqref{two_bar_problem} suffers from singular optima, which are solutions (in this case the global optima) existing in lower dimensional extensions to the feasible design space. Accessing such optima using gradient-based optimizers is extremely difficult unless some form of constraint relaxation is applied.

As shown in Fig.~\hyperref[fig:stress2d_qpm]{\ref*{fig:stress2d_qpm}a}, the global minimum can only be accessed through the one-dimensional extension of the feasible domain (shown hashed). Note that each constraint has two branches, one corresponding to the sloped line and another to the coordinate axis itself (a consequence of the vanishing constraint).
Relaxation methods such as $\epsilon$- relaxation~\cite{Cheng1997} or \textit{qp}-relaxation~\cite{Bruggi2008}, create a larger access zone by allowing the constraint to be of the form $\bar{g}_i \leq \epsilon$, where $\epsilon$ is a small scalar; the latter is gradually reduced to zero so that optimization converges to the original problem. After relaxation, the global minimum becomes accessible and optimizers like MMA~\cite{Svanberg1987} may converge to the global optimum.

To compute hypergradients, we prefer first-order unconstrained 
optimizers such as Adam over constrained optimizers like MMA: AD through Adam's simple update rule 
is numerically more stable than through MMA's constrained subproblem solves. We therefore convert the constrained optimization to an 
unconstrained one using the \textit{quadratic penalty method} (QPM)\footnote{The squared-penalty form 
$\bar{g}_i^2$ corresponds to equality constraints; inequality 
constraints would use $[\max(0, \bar{g}_i)]^2$. We adopt the equality 
form here because both constraints are active at the local and global 
optima, so they behave as equalities.}:
\begin{equation}\label{eqn:augobj}
    \boldsymbol{a}^\star = \lim_{\mu \to \infty} \, 
    \argmin_{\boldsymbol{a} \in [0, 2]^2} \; \Big[ Q(\bs{a};\mu) = f(\boldsymbol{a}) + 
    \frac{\mu}{2} \sum_i \bar{g}_i(\boldsymbol{a})^2 \Big].
\end{equation}
where the augmented objective ($Q$) adds a penalty ($\mu$) for constraint 
violations to the original objective $f$. As $\mu \to \infty$, the 
minimizer of $Q$ recovers a KKT point of the original constrained
problem~\eqref{two_bar_problem}~\cite{nocedal2006numerical}. In practice, QPM is solved iteratively. Starting from a small penalty 
$\mu^0$, the augmented objective is approximately minimized over the 
design variables $\boldsymbol{a}$; the penalty is then increased and 
the minimization repeated, continuing until $\mu$ is sufficiently 
large or the constraints are satisfied to a prescribed tolerance. 
Small $\mu$ keeps $Q$ smooth and lets the optimizer explore broadly 
while remaining infeasible with respect to the true constraints; as 
$\mu$ grows, the penalty progressively enforces feasibility (we keep an upper bound of $10^{4}$ for numerical stability). 
Fig.~\hyperref[fig:stress2d_qpm]{\ref*{fig:stress2d_qpm}b} shows the 
augmented objective for several values of $\mu$.

We observe that each QPM subproblem, $\min_{\boldsymbol{a}} 
Q(\boldsymbol{a}; \mu^{(k)})$, can be reformulated as a relaxed 
constrained problem closely analogous to existing $\epsilon$-relaxation 
methods. Whereas $\epsilon$-relaxation adds and controls an explicit 
scalar $\epsilon$ to open up the feasible region, QPM produces this 
relaxation implicitly. We relate the two formulations in 
Appendix~\ref{app:qpm_eps_relax} and show that, at the exact 
subproblem minimizer, the constraint is relaxed by an amount
\footnote{Here $\lambda_i^\star$ is the Lagrange multiplier of the 
$i$-th constraint in the original constrained 
problem~\eqref{two_bar_problem}, obtained by solving its \textit{Karush-Kuhn-Tucker} (KKT) 
conditions. Because the problem is non-convex with two KKT points 
(a local and a global minimum), each admits its own multiplier 
vector, and the relaxation is therefore dependent 
on the basin toward which QPM converges.}
$\epsilon^{(k)}_i = \lambda_i^\star / \mu^{(k)}$. 
Fig.~\hyperref[fig:stress2d_qpm]{\ref*{fig:stress2d_qpm}c} shows the 
relaxed constraint boundaries at two values of $\mu$, using the 
global optimum's multipliers $\boldsymbol{\lambda}^\star = (1.2, 0.4)$ (obtained analytically). 
As $\mu$ increases, the relaxed boundaries contract toward the true 
constraints, equivalent to taking $\epsilon \to 0$ in 
$\epsilon$-relaxation.

\begin{figure*}
    \centering
    \includegraphics[width=1\linewidth]{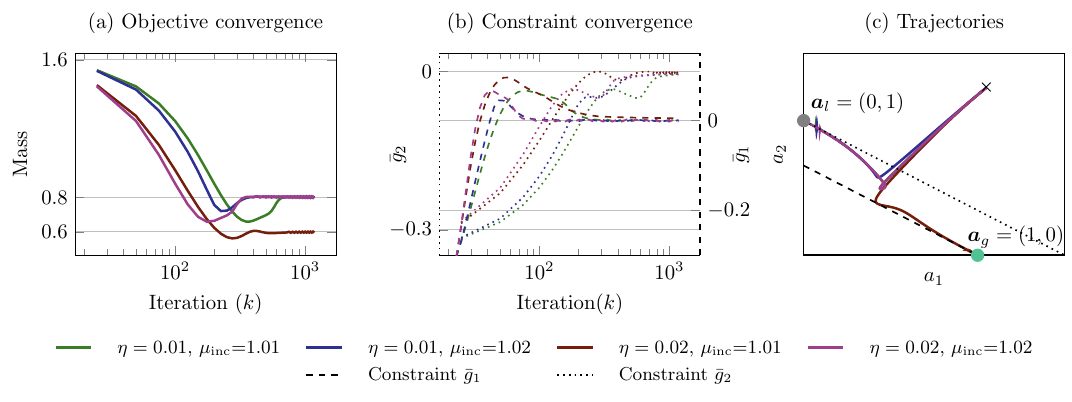}
\caption{Sensitivity of the quadratic penalty method (QPM) to hyperparameters on the benchmark two-bar stress-constrained optimization problem. The original 
constrained problem is converted via QPM into an unconstrained 
formulation and solved by Adam, with the penalty $\mu$ incremented following $\mu^{(k)} =\mu_{\text{inc}}\mu^{(k-1)}$. 
Four trajectories are shown, corresponding to two values of $\mu_{\text{inc}}$ paired with two learning rates $\eta =  \left \{0.01, 0.02 \right\}$. 
(a) Convergence of the objective: only one trajectory reaches the 
globally optimum $f \left(\bs{a}_\star \right) = 0.6$; 
(b) Evolution of the constraints for each of the runs, with feasible values $\bar{g}_i = 0$ (One constraint with dashes and other with dots); 
(c) Trajectories in the design space $\boldsymbol{a}$, with constraint boundaries shown in black lines.}
\label{fig:qpm_sensitivity}
\end{figure*}

However, which minimum QPM converges to depends strongly on the 
hyperparameters. Fig.~\ref{fig:qpm_sensitivity} demonstrates this for 
the stress-constrained problem. We increase the penalty 
multiplicatively as $\mu^{(k)} = \mu_{\text{inc}} \, \mu^{(k-1)}$, and 
consider two growth factors $\mu_{\text{inc}} \in \{1.01, 1.02\}$ and 
two Adam learning rates $\eta \in \{0.01, 0.02\}$. Starting from 
$\boldsymbol{a} = (1.2, 1.2)$, we run all four combinations, 
incrementing the penalty after each single LL step. Only one of the four trajectories 
converges to the global optimum, and which one does depends 
sensitively on the joint interaction between the penalty growth and 
the step size. This is exactly the kind of multi-hyperparameter sensitivity that BOTH is designed to resolve.

Building on the QPM-$\epsilon$ relaxation equivalence and the hyperparameter sensitivity it inherits, we reformulate the constrained problem~\eqref{two_bar_problem} as a bilevel optimization problem with two 
UL hyperparameters (the QPM penalty $\mu$ and Adam's learning rate $\eta$):
\begin{equation} \label{eq:bilevel_2d_stress}
\begin{aligned}
    \min_{\boldsymbol{y} = \{\mu, \eta\}} \;\; & 
    F(\boldsymbol{x}_\star) = \frac{f(\boldsymbol{x}_\star)}{f^u_{\text{ref}}} + 
    k \sum_i \frac{\bar{g}_i(\boldsymbol{x}_\star)^2}{p^u_{\text{ref}}} \\
    \text{such that} \quad \boldsymbol{x}_\star &= 
    \argmin_{\boldsymbol{x}} \frac{f(\boldsymbol{x})}{f^l_{\text{ref}}} + 
    \mu \sum_i \frac{\bar{g}_i(\boldsymbol{x})^2}{p^l_{\text{ref}}},
\end{aligned}
\end{equation}
%
\noindent where the reference values $f_{\text{ref}}$ and $p_{\text{ref}}$ are
chosen so that the objective and penalty terms are of unit magnitude at
initialization. The same normalization is applied independently at both levels,
with the reference values evaluated from the respective level's own objective and
penalty terms. The LL design variables $\boldsymbol{x}$ are mapped to bar areas 
$\boldsymbol{a} = h(\boldsymbol{x})$ via a parameterization that 
enforces the box bounds $a_i \in \left[ 0, a_{\max} \right]$. For the standard 
parameterization used here, $h_i = a_{\max} s(x_i)$, where 
$y = s(x)$ is the sigmoid function, giving two LL variables for our two-bar problem.

The factor $k$ in the UL objective is an initialization-dependent weight, and its
role is best understood through a degeneracy in the limiting case. Unlike the
compliance example in Section~\ref{subsec:compliance_bilevel}, the UL and LL
objectives here share the same functional form and differ only in the penalty
weight ($k$ at the UL, $\mu$ at the LL). If $k = \mu$ the two are identical
functions of $\boldsymbol{x}$, so the LL optimum $\boldsymbol{x}_\star$ is also a
stationary point of the UL objective; the hypergradient, which flows entirely
through $\boldsymbol{x}_\star$, then vanishes because
$\partial F / \partial \boldsymbol{x}_\star = \boldsymbol{0}$. The separation
$k \neq \mu$ is therefore what generates hypergradient signal at all, and the size
of the separation sets how strongly the UL drives $\mu$ upward.

This makes $k$ a target rather than a free parameter. Under QPM, the
constraint is satisfied exactly only as $\mu \to \infty$, so the design is always
slightly infeasible and the meaningful specification is a tolerance on satisfying the constraints rather than feasibility itself. We therefore set $k$ to an estimate of the penalty
multiplier at which the worst-violating element would reach a prescribed tolerance
on the stress margin (we set $\texttt{tol} = 10^{-3}$) at the end of optimization. The estimate reduces to
\begin{equation}
    k = \frac{|g_b|}{\texttt{tol}},
    \label{eqn:ul_factor}
\end{equation}
\noindent where $g_e = \sigma_e/\sigma_{\max} - 1$ is the raw stress margin and $b$
is the worst-violating element at the initial design variables. The tolerance thus bounds the
physical stress margin directly, and the estimate requires only quantities available
at initialization, with no unrolling. The general form, its derivation, and the
conditions under which it collapses to Eqn.~\eqref{eqn:ul_factor} are given in
Appendix~\ref{app:impl_loss}.

\begin{figure*}[t]
    \centering
    \includegraphics[width=1.0\textwidth]{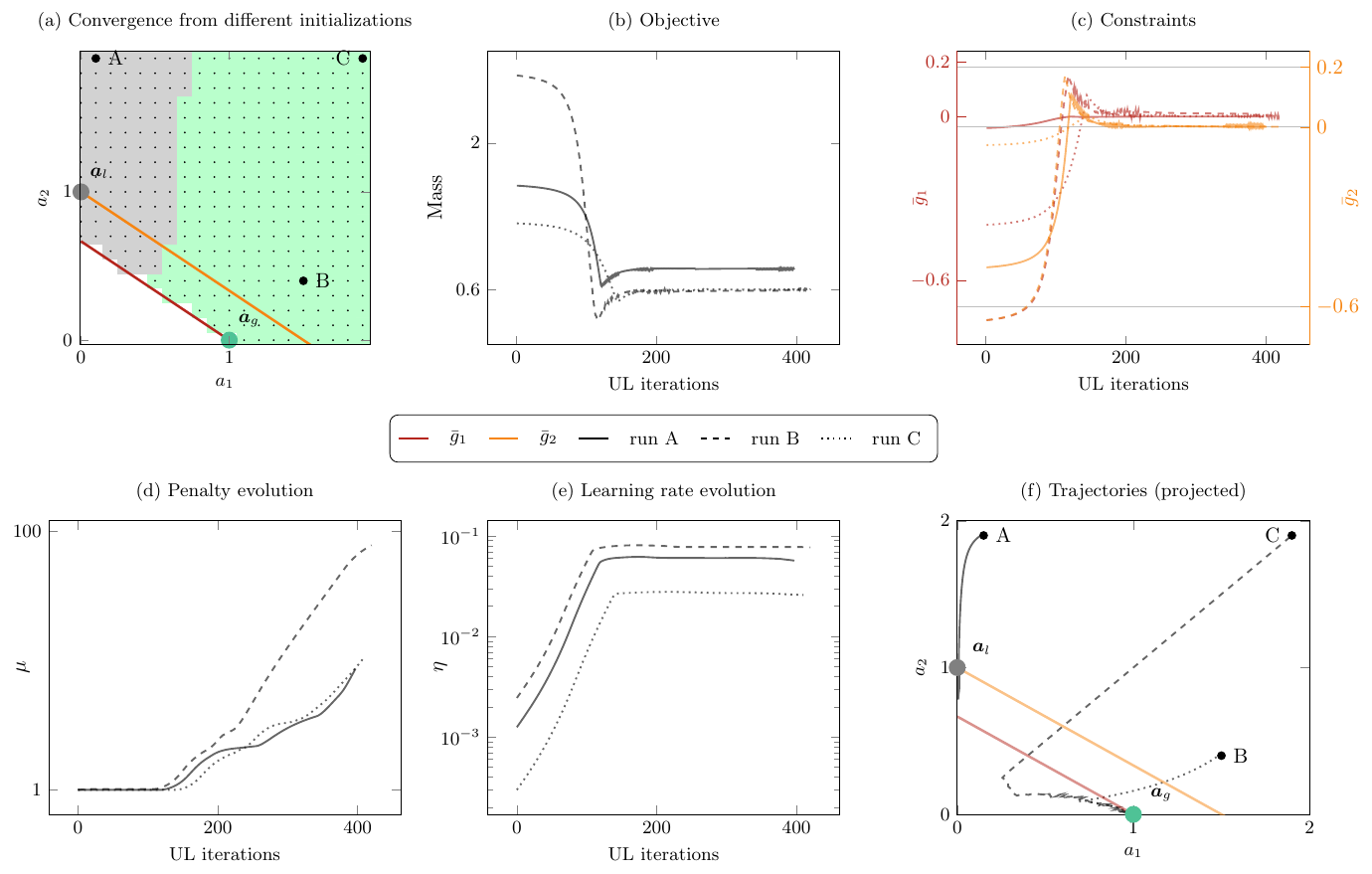}
\caption{Results for the two-dimensional stress problem with standard 
parameterization. (a) Each dot represents one optimization run from a 
distinct LL initialization; green indicates convergence to the global 
optimum, gray indicates convergence to the local minimum. 
(b, c) Convergence histories of the objective and the two constraints 
for three representative runs (A, B, C). (d, e) Learned schedules for 
the LL learning rate $\eta$ and the QPM penalty $\mu$ (both 
parameterized in log scale) along these three runs. (f) Trajectories 
projected into the $(a_1, a_2)$ space.}
    \label{fig:standard}
\end{figure*}


The results for the standard parameterization are shown in 
Fig.~\ref{fig:standard}, where we run the bilevel optimization from 
several starting points. Both $\mu$ and $\eta$ are reparameterized in log 
space (e.g., $\mu = 10^{y_1}$), since they span orders of magnitude, 
and $\mu$ is constrained to be non-decreasing, in line with standard QPM convergence requirements~\cite{nocedal2006numerical}. The black dots in 
Fig.~\hyperref[fig:standard]{\ref*{fig:standard}a} show the initial 
LL variables for each run, uniformly spread across the design space. 
For each run, we tune the LL learning rate and the QPM penalty using 
our framework. The square surrounding each dot indicates the outcome: 
green for convergence to the global optimum, gray for entrapment in 
the local optimum. The broad spread of green squares across the 
design space empirically confirms the relaxation effect of QPM: 
without any explicit constraint relaxation, many runs nonetheless 
reach the global optimum.

Three runs are marked A, B, and C in 
Fig.~\hyperref[fig:standard]{\ref*{fig:standard}a}: B and C converge 
to the global optimum, while A becomes trapped in the local optimum. 
The objective and the two constraints 
(Figs.~\hyperref[fig:standard]{\ref*{fig:standard}b} 
and~\hyperref[fig:standard]{\ref*{fig:standard}c}) show the 
constraints being satisfied and the mass settling to $0.6$ (global) 
or $0.8$ (local), confirming convergence to a KKT point of the 
constrained problem. Because we use QPM, exact feasibility is never 
reached; the constraints are met only up to a tolerance that 
tightens as $\mu$ increases. Oscillations can appear as the augmented 
landscape becomes increasingly ill-conditioned at large $\mu$ 
(Fig.~\hyperref[fig:stress2d_qpm]{\ref*{fig:stress2d_qpm}b}).

The learned schedules for $\mu$ and $\eta$ 
(Figs.~\hyperref[fig:standard]{\ref*{fig:standard}d} 
and~\hyperref[fig:standard]{\ref*{fig:standard}e}) are qualitatively 
similar across the three runs but differ in detail---for example, 
$\mu$ stays near its initial value for the first few iterations in 
all runs but subsequently grows at different rates. Projecting the 
trajectories into the $(\boldsymbol{a}_1, \boldsymbol{a}_2)$ space 
(distinct from the design space, since a sigmoidal transform is 
applied) clarifies the outcomes 
(Fig.~\hyperref[fig:standard]{\ref*{fig:standard}f}). Run A overshoots 
the local minimum early, when the penalty is small and the local 
basin is shallow; as the penalty grows, that basin deepens and traps 
A. Runs B and C instead follow penalty schedules that keep the global 
minimum's basin accessible long enough for the trajectory to settle 
there before the penalty becomes large.




\section{Bilevel optimization applied to the benchmark L-shaped bracket problem}

\begin{figure}
    \centering
    \includegraphics[width=1\columnwidth]{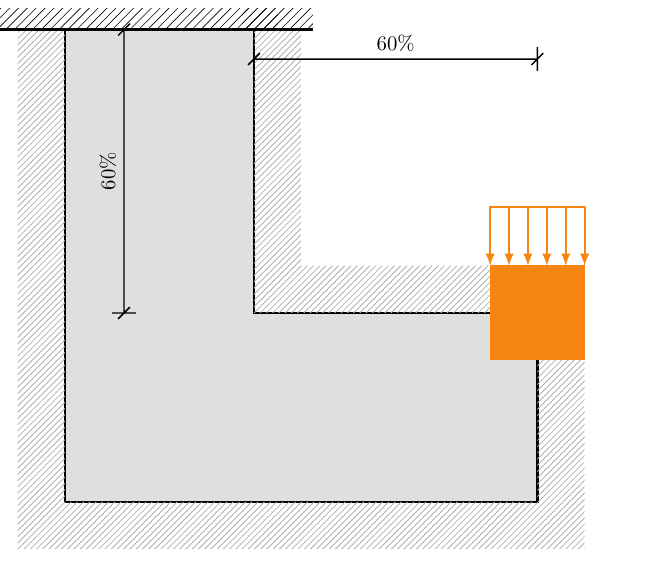}
    \caption{Boundary conditions for the L-shaped bracket problem. The top edge is fully fixed, and a downward force is distributed over a 
non-designable solid region (orange square). A padding layer (hatched) of non-designable void elements surrounds the domain and is included 
in the FEA, so that internal and external boundaries are treated consistently~\cite{amir_stress}. The designable elements lie within the L-shaped domain.}
    \label{fig:lshapebc}
\end{figure}

\begin{figure*}[t]
    \centering    \includegraphics[width=1.0\textwidth]{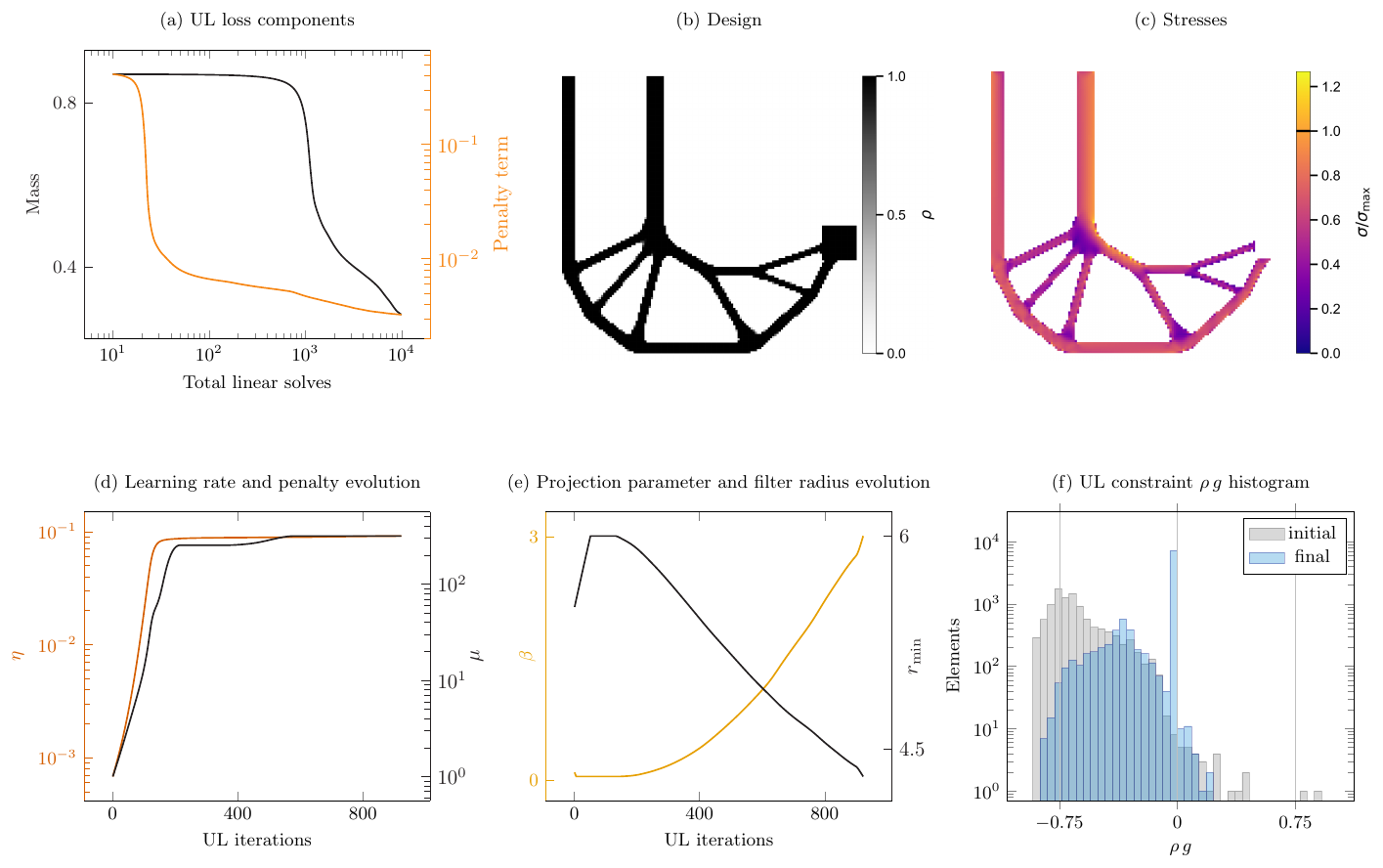}
    \caption{Results for tuning hyperparameters in the design of an L-shaped beam under
    stress constraints. (a) Convergence of the two components of the upper-level loss,
    the objective (mass) and the penalty term (sum of squared constraints),
    against the number of linear solves consumed by the bilevel optimization. (b) Final
    design, projected with $\beta = 64$. (c) Ratio of von Mises stress $\sigma$ to the
    stress limit $\sigma_{\max}$ on elements with density above $0.5$, evaluated
    assuming solid material. (d) Evolution of the lower-level learning rate and of the
    penalty factor $\mu$ of the quadratic penalty method. (e) Evolution of the
    projection parameter $\beta$ and the radius of the density filter. (f) Distribution of the upper-level constraint values
    at the start and end of the bilevel optimization.}
    \label{fig:l_shape}
\end{figure*}
The L-shaped bracket problem (Fig.~\ref{l_shape_problem}) is a widely used benchmark for continuum stress TO and remains challenging because of its pronounced sensitivity to the choice of hyperparameters~\cite{amir_stress}. The L-shaped design domain has its top-most edge fixed and a downward force applied at the right-most edge. This choice of boundary conditions results in a stress concentration at the re-entrant corner. Therefore, the expected topology is one that would smoothen that region. The problem is formulated as mass minimization subject to a stress constraint on each element:
\begin{equation} \label{l_shape_problem}
	\begin{aligned}
		\bs{\rho}_{\star} = \argmin_{ \bs{\rho} \in \mathbb{R}^N} \quad &  f\left( \bs{\rho}\right) = \frac{\sum_i^N \rho_i v_i}{\sum_i^N v_i}, \\
		\text{such that} \quad & \mathbf{K}(\bs{\rho})\bs{u} = \bs{f}\\
        & \bar{g}_i \leq 0, \\
		\quad & 0 \leq \rho_i \leq 1, \quad i \in \left\{ 0, \ldots, N \right\}, \\
	\end{aligned}
\end{equation}
where $N$ is the number of designable elements, $\bs{\rho}$ the physical densities, $\bar{g}_i$ the vanishing constraints, and $v_i$ element volumes. As in the compliance problem 
(Eqn.~\ref{eq:compliance_LL}), we use the SIMP law for 
stiffness interpolation (with $p = 3$) and the same density and projection filters. 
The linear elastic equilibrium equation $\mathbf{K}\boldsymbol{u} = \boldsymbol{f}$ 
is solved with a direct solver; further FEA details are in Appendix~\ref{app:lshape}. 

Conventionally, the large number of local stress constraints is 
aggregated into a single constraint via a $p$-norm or 
Kreisselmeier--Steinhauser (KS) function, combined with a relaxation 
scheme\footnote{\cite{Verbart2016} show that modifying these 
functions can eliminate the need for relaxation.}, and solved with 
MMA using adjoint sensitivities. As in the two-bar problem 
(Section~\ref{sec:LL_constraints}), we instead treat each constraint 
locally---without aggregation---using QPM with Adam as the 
unconstrained optimizer~\cite{Senhora2020, silva_stress}. 

Relative to the two-bar problem, the L-shape introduces two difficulties. First, the number of constraints is large and grows 
with mesh resolution, rather than being fixed at two. Second, the 
stiffness and the stress measure are treated inconsistently: 
stiffness follows the SIMP interpolation, while the von Mises stress 
$\sigma_i$ at each element centroid is computed assuming solid 
material ($E_0 = 1$). We define the constraint as $g_i = \sigma_i / 
\sigma_{\max} - 1$ with allowable stress $\sigma_{\max} = 0.5$.  To address the latter, we adopt the density-weighted constraint form of 
\cite{Senhora2020}, $\bar{g}_i = \rho_i^p g_i^2$, at the LL, where $p$ is the SIMP penalty factor. The 
standard vanishing form caused QPM to ignore the re-entrant-corner 
constraints entirely, yielding a design without the expected 
rounding; the difficulty is compounded by the sigmoid projection, 
which makes it hard to drive densities fully to zero, so near-void 
elements still contribute small nonzero constraint terms.

At the UL, we adopt the standard vanishing constraint. In addition to 
the penalty $\mu$ and learning rate $\eta$, we tune the projection 
sharpness $\beta$ and the density filter radius $r_{\min}$. For these latter two hyperparameters, the UL objective is evaluated using the same values employed at the LL, consistent with the compliance problem. All other settings match the 
previous experiments; the LL is initialized with all element densities close to $0.9$, following standard 
practice~\cite{amir_stress}. The bilevel formulation is:
\begin{equation} \label{eq:bilevel_lshape}
\begin{aligned}
    \min_{\boldsymbol{y} = \{\mu, \eta, \beta, r_{\min}\}} \;\; & 
    F(\boldsymbol{x}_\star) = \frac{f(\boldsymbol{x}_\star)}{f^u_{\text{ref}}} + 
    \frac{k}{N} \sum_i \frac{\left(\rho_i{g}_i(\boldsymbol{x}_\star)\right)^2}{p^u_{\text{ref}}} \\
    \text{s.t.} \quad \boldsymbol{x}_\star &= 
\argmin_{\boldsymbol{x}} \; \frac{f(\boldsymbol{x})}{f^l_{\text{ref}}} + 
    \frac{\mu}{N} \sum_i \frac{\left(\rho_i^p{g}^2_i(\boldsymbol{x})\right)^2}{p^l_{\text{ref}}}.
\end{aligned}
\end{equation}
\noindent where the reference values $f_{\text{ref}}$ and $p_{\text{ref}}$ are set
independently at each level, so that at initialization, the two terms of that level's
objective are equal in magnitude, following the same approach as for the two-bar
problem. Note that the modified
constraint appears only at the LL. The factor $k$ is set by the same
strategy as in the two-bar problem, though the resulting formula differs (Appendix~\ref{app:impl_loss}) because the
two levels no longer share the same loss form.

The results of the bilevel optimization of the L-bracket are shown in
Fig.~\ref{fig:l_shape}. Fig.~\hyperref[fig:l_shape]{\ref*{fig:l_shape}a} shows the
evolution of the two components of the UL objective, the mass and the
penalty term. Both decrease as the optimization proceeds, and neither has converged
at the point at which the run was stopped by the allotted budget. The design at that
stage was correspondingly not fully black-and-white, so we project it at a higher
$\beta = 64$; the result, shown in
Fig.~\hyperref[fig:l_shape]{\ref*{fig:l_shape}b}, exhibits a rounded re-entrant
corner and no structural disconnections.

For this nearly binary design, the ratio of von Mises stress to the allowable stress
is shown in Fig.~\hyperref[fig:l_shape]{\ref*{fig:l_shape}c} for elements with
$\rho_i > 0.5$. The stresses lie close to the allowable value across the structure,
although the maximum remains above the limit and the design is therefore still
infeasible---an outcome expected of the QPM, under which the
constraint is approached only as the penalty factor grows without bound. The
hyperparameter trajectories in
Fig.~\hyperref[fig:l_shape]{\ref*{fig:l_shape}d} are consistent with this: the
penalty factor rises from $1.0$ to approximately $500$, a value at which a residual
violation is still expected, while the learning rate increases throughout and
stagnates near its upper bound of $0.1$, having started two orders of magnitude
below it. The projection parameter likewise increases
(Fig.~\hyperref[fig:l_shape]{\ref*{fig:l_shape}e}), promoting binary designs
particularly in the later stages of the optimization.
The evolution of the filter radius differs notably from the compliance problem: rather than decreasing monotonically, it rises early in the optimization before subsequently falling. Comparing the initial and
final distributions of the UL constraint over all elements
(Fig.~\hyperref[fig:l_shape]{\ref*{fig:l_shape}f}) shows that the magnitude of the
violations has decreased and that many elements have moved towards a fully stressed
state, as the equality form of the constraint intends.

\section{Conclusion}
We have introduced a bilevel framework for optimizing hyperparameters and design parameters within a single optimization run, deriving hypergradients by differentiating through TO. A central empirical finding of this work is that the resulting gradients are usable after unrolling only one or two LL steps. Hyperparameter
landscapes in TO are highly non-convex, and a hypergradient taken
through such a short trajectory has no guarantee of being informative. We find
nonetheless that the warmstarting strategy, together with annealing the number of inner steps
when the UL stagnates, smooths the unrolled trajectory sufficiently for the
hypergradient to carry reliable signal. We demonstrated the methodology on both stress-constrained TO and on compliance minimization problems, the latter with a neural parameterization of the density field. Because the cost of an unrolled hypergradient
is dominated by the linear solves involved, and the number of those solves is set
by the unroll length rather than by the number of hyperparameters, the framework
scales naturally to high-dimensional settings such as per-element fields comprising
thousands of variables. Even on a problem with four hyperparameters, where Bayesian
optimization is at its most competitive, it matches or slightly exceeds the quality
of the designs obtained under similar compute budgets.

The framework is nevertheless nascent and faces several limitations. It is
restricted to continuous hyperparameters; discrete algorithmic choices still require
continuous relaxations or non-gradient methods. It is also invasive, since
differentiating through the optimization trajectory requires an
AD-friendly implementation of the entire nested procedure.
The search is greedy and explores the region near its starting point, so the initial
hyperparameter values remain a design decision in their own right, alongside the two new hyperparameters at the UL i.e., choosing the factor $k$ and UL optimizer's learning rate. A
further difficulty is that the UL can satisfy its objective by loosening the
problem rather than by improving the design: were the allowable stress $\sigma_{\max}$
exposed as an unrestricted hyperparameter, for instance, the UL would simply
raise it, reducing the penalty without producing a better structure. Admissible
hyperparameter channels must therefore be restricted by construction, and these
restrictions are presently designed by hand for each formulation. More generally, the framework does not eliminate user judgment but relocates it from the choice of individual hyperparameter values to the definition of the bilevel formulation, the normalization of the two objectives, and the initialization heuristics that precede
the run.

Several directions follow. To make the framework less invasive and more readily
transferable across problems, a hypergradient estimate based on the implicit function
theorem would be preferable. Such estimates capture only those hyperparameters that move the LL optimum. Recovering signal for trajectory hyperparameters therefore requires modifying the bilevel formulation itself, ensuring these quantities enter the stationarity condition rather than acting only on the path taken; the proximal
formulation of implicit MAML~\cite{rajeswaran2019meta} is one instance of this
strategy. Surrogate models offer another route to reducing cost, and their own
training settings could in turn be exposed as hyperparameters. The treatment of
constraints at the LL also merits revisiting: because the quadratic penalty
aggregates over elements, the LL can tolerate a few severe violators in
exchange for reducing the penalty across the bulk of the domain, an imbalance that
the equality form of the constraint does not discourage. Replacing the LL
optimizer with MMA would enforce the constraints directly rather than through a
penalty, and would expose its own asymptote parameters as hyperparameters in turn.
The set of admissible channels is likewise far from exhausted: per-element filter
radii, per-parameter learning rates, and preconditioners for the linear solves are
all natural candidates. In principle any continuous setting in the nested
procedure---in the topology optimization, in the finite element analysis, or in the
involved solvers---is admissible, provided the operations concerned are smooth enough
for the hypergradient to carry signal. Extending the framework to discrete choices
would widen this scope further, bringing neural architecture itself within reach
alongside the continuous parameterization hyperparameters already handled.

The broader implication is that the algorithmic ingredients of TO
need not remain fixed, hand-crafted choices. Penalization factors, projection
parameters, aggregation parameters, relaxation parameters, move limits, and
continuation schedules are typically selected through accumulated experience and
problem-specific trial and error. Treating these quantities as differentiable
variables opens a path toward topology optimization procedures whose algorithmic
policies are optimized for the problem class at hand. It also removes a standing reason to keep hyperparameter spaces small. Formulations that would previously have been impractical to tune---because they expose spatially varying or per-parameter settings---become accessible once those settings can be optimized to the problem at hand rather than chosen apriori.

\section*{Acknowledgments}{
All authors sincerely thank Prof. Fred van Keulen, Prof. Mathijs Langelaar, and the whole Computational Design and Mechanics group from the Mechanical Engineering faculty at TU Delft for the fruitful discussions and valuable feedback provided. S.M.S. appreciates the fruitful discussions regarding bilevel optimization with Prof. Ankur Sinha (Indian Institute of Management Ahmadabad).}

\newpage

\appendix

\section{Implementation details of BOTH} \label{app:hyperto_details}

The implementation details corresponding to the components of 
Algorithm~\ref{alg:hyperto} are described below. The entire framework, 
including the topology optimization routines, is implemented in 
JAX~\cite{jax2018github}, which provides automatic differentiation 
infrastructure for computing the unrolled hypergradient through the 
LL trajectory.
 
\subsection{Computational budget allocation}
\label{app:impl_budget}
 
The user specifies a total computational budget $\mathcal{B}_{\max}$ in cumulative 
LL iterations. This budget is distributed across the annealing stages 
$t = 1, 2, \ldots, t_{\max}$ according to an exponentially decaying 
schedule with decay factor $0.1$, biasing 
allocation toward early stages where each UL update requires fewer LL 
evaluations. Two hard reserves are imposed regardless of convergence 
behavior: $20\%$ of $\mathcal{B}_{\max}$ is reserved for $t = 1$ and 
$10\%$ for $t = t_{\max}$. These floors ensure sufficient exploration 
in the cheap early regime ($t = 1$) and adequate optimality enforcement 
in the final regime ($t = t_{\max}$).
 
Within each stage, convergence checks (Appendix~\ref{app:impl_convergence}) 
are disabled during the first $10\%$ of that stage's allocated budget. 
This \textit{patience} window prevents premature stage transitions caused by 
transient behavior immediately following a change in $t$.
 
\subsection{Convergence and annealing triggers}
\label{app:impl_convergence}
 
A transition from stage $t$ to stage $t+1$ is triggered when 
optimization at the current stage has stagnated. Stagnation is detected 
through two quantities: the 
relative change in the UL objective $F$ and the relative change in the 
UL variables $\boldsymbol{y}$, defined as
\begin{equation}
    \Delta_F^{(k)} = \frac{|F^{(k)} - F^{(k-1)}|}{|F^{(k-1)}|}, 
    \quad 
    \Delta_{\boldsymbol{y}}^{(k)} = 
    \frac{\|\boldsymbol{y}^{(k)} - \boldsymbol{y}^{(k-1)}\|}
         {\|\boldsymbol{y}^{(k-1)}\|}.
\end{equation}
Rather than using instantaneous values, , both quantities are tracked 
via their exponentially weighted moving averages (EMA),
\begin{equation}
    s^{(k)} = 0.9\, s^{(k-1)} + 0.1\, \Delta^{(k)}, 
    \quad s^{(0)} = \Delta^{(0)},
\end{equation}
where $\Delta^{(k)}$ is either $\Delta_F^{(k)}$ or 
$\Delta_{\boldsymbol{y}}^{(k)}$.
The smoothing factor $0.1$ 
corresponds to an effective window of approximately ten UL iterations. 
A plateau transition is triggered when both EMA-smoothed quantities 
fall below $10^{-3}$.

In addition, oscillation is monitored via the ratio of the EMA of 
the signed loss change to the EMA of the absolute loss change. Formally, the oscillation ratio at iteration $k$ is
\begin{equation}
    r^{(k)} = \frac{\text{EMA}(F^{(k)} - F^{(k-1)})}
                   {\text{EMA}(|F^{(k)} - F^{(k-1)}|)},
\end{equation}
where both EMAs use a smoothing factor of $0.1$, consistent with the 
plateau check above. A ratio $r^{(k)} < 0.5$ indicates that signed 
and absolute changes are of comparable magnitude, i.e., progress is 
made in fewer than half of recent iterations. If this 
condition persists for $30$ consecutive UL iterations, the stage is 
incremented regardless of whether the plateau criterion is met. EMAs 
are carried across stage transitions, except for the oscillation 
tracking EMA, which is reset at each transition.

A forced transition to $t = t_{\max}$ is triggered when the remaining 
budget $\mathcal{B}_{\max} - \mathcal{B}$ equals the reserve allocated 
to $t = t_{\max}$, ensuring that the final stage receives its full 
allotment.

\subsection{Estimating UL factor}
\label{app:impl_loss}

We derive the general form of Eqn.~\eqref{eqn:ul_factor}, state the
conditions under which it collapses to the expression used in the main text, and
records the assumptions that the estimate rests on. This is applicable for the two stress constrained examples that we consider where, both levels minimize an objective of the form
\begin{equation}
	\frac{m}{m_{\text{ref}}} \;+\; w\,\frac{\operatorname{op}\!\left(\boldsymbol{h}^2\right)}{p_{\text{ref}}},
	\label{eqn:app_common_form}
\end{equation}
\noindent with penalty weight $w = \mu$ at the lower level and $w = k$ at the
upper, and with level-specific normalization scales $(m^l_{\text{ref}}, p^l_{\text{ref}})$ and $(m^u_{\text{ref}}, p^u_{\text{ref}})$. Here $m$ is the mass, $\operatorname{op}$ is the
aggregation used by the constraint form, and
\begin{equation}
	h_e \;=\; \rho_e^{\,p}\, g_e^{\,q},
	\qquad
	g_e \;=\; \frac{\sigma_e}{\sigma_{\max}} - 1,
\end{equation}
\noindent is the penalized quantity, with $g_e$ the raw stress margin at element
$e$ and $(p,q)$ fixed by the constraint form. The two-bar problem uses $p=q=1$ and the L-bracket problem sets $p=3,q=2$. The derivation proceeds in three
steps: extracting an effective multiplier from the lower level, converting a
tolerance on $g$ into a target multiplier, and converting that LL weight
into an UL one.

\paragraph{Step 1: effective multiplier.}
Comparing the stationarity condition of the loss the LL actually minimizes (gradient of the augmented objective is zero at the minimizer),
\begin{equation}
	\frac{\nabla m}{m^l_{\text{ref}}}
	+ \frac{2\mu}{p^l_{\text{ref}}}\sum_e h_e \nabla h_e \;=\; \boldsymbol{0},
\end{equation}
\noindent against the KKT condition of the unnormalized constrained problem,
$\nabla m + \sum_e \lambda_e \nabla h_e = \boldsymbol{0}$, gives
\begin{equation}
	\lambda_e \;=\; 2\mu_e \frac{m^l_{\text{ref}}}{p^l_{\text{ref}}}\, h_e .
	\label{eqn:app_multiplier}
\end{equation}
\noindent This identity is exact with respect to the objective the LL is
optimizing, since the normalization scales are part of tne that objective (fixed at initialization). Two
consequences follow. First, the residual violation at an LL stationary point is
$h_e = (p^l_{\text{ref}}/2m^l_{\text{ref}})\,\lambda_e/\mu_e$, which vanishes only as
$\mu \to \infty$: under a quadratic penalty the constraint is approached but never
met, which is what makes a tolerance the appropriate specification. Second, the LL
scales appear inside $\lambda$ and therefore never need to be retrieved separately.

\paragraph{Step 2: tolerance conversion.}
Let $b = \arg\max_e |\lambda_e| = \arg\max_e |\mu_e h_e|$ denote the binding
element; by Eqn.~\eqref{eqn:app_multiplier} the two definitions coincide, and the
form using $\mu_e h_e$ accommodates a per-element multiplier field without special
handling (we use $\mu_e = \mu$ for all elements). Requiring the margin at  the element $b$ to satisfy $|g_b| \leq \texttt{tol}$ and
converting through the same constraint form at the same element,
$h_{\text{target}} = \rho_b^{\,p}\,\texttt{tol}^{\,q}$, yields
\begin{equation}
	\mu_{\text{target}}
	\;=\; \mu_b \frac{|h_b|}{h_{\text{target}}}
	\;=\; \mu_b \frac{\rho_b^{\,p}\,|g_b|^{\,q}}{\rho_b^{\,p}\,\texttt{tol}^{\,q}}
	\;=\; \mu_b \left(\frac{|g_b|}{\texttt{tol}}\right)^{q}.
	\label{eqn:app_mu_target}
\end{equation}
\noindent The density factor $\rho_b^{\,p}$ cancels identically between the current
and target values. This is the reason the tolerance is specified on $g$ rather than
on $h$: \texttt{tol} then bounds the physical stress margin, and the constraint form
enters only through the exponent $q$.

\paragraph{Step 3: lower- to upper-level conversion.}
Eqn.~\eqref{eqn:app_mu_target} is a target specification for the LL weight, whereas $k$ is associated to the UL. The two are linked by the UL stationarity condition
$M + k P = 0$, where $M = \nabla m \cdot \mathbf{J}/m^u_{\text{ref}}$,
$P = \nabla p \cdot \mathbf{J}/p^u_{\text{ref}}$, and
$\mathbf{J} = \mathrm{d}\boldsymbol{x}_t/\mathrm{d}\mu$ is the sensitivity
of the LL terminal state to the multiplier. Substituting LL stationarity and
writing each penalty-gradient projection as
$\nabla p \cdot \mathbf{J} = \lVert\nabla p\rVert\,\lVert\mathbf{J}\rVert\cos\theta$
gives $k = \mu_{\text{target}}\, C$ with
\begin{equation}
	C \;=\;
	\underbrace{\frac{m^l_{\text{ref}}\,p^u_{\text{ref}}\,\lVert\nabla p_l\rVert}
		{m^u_{\text{ref}}\,p^l_{\text{ref}}\,\lVert\nabla p_u\rVert}}_{C_{\text{known}}}
	\;\cdot\;
	\underbrace{\frac{\cos\theta_l}{\cos\theta_u}}_{C_{\text{align}}} .
	\label{eqn:app_C}
\end{equation}
\noindent Collecting Eqns.~\eqref{eqn:app_mu_target} and~\eqref{eqn:app_C},
\begin{equation}
	k \;=\; \mu_b \left(\frac{|g_b|}{\texttt{tol}}\right)^{q} C_{\text{known}}\, C_{\text{align}} .
	\label{eqn:app_ul_factor_general}
\end{equation}

The factor $C_{\text{known}}$ requires only the four normalization scales and the
two penalty-gradient norms, all available at initialization without unrolling. The factor $C_{\text{align}}$ is taken as unity.  For the stress-constrained problems in this work, $q = 1$, both levels are
conditioned on the same mass term so $C_{\text{known}} = 1$, both penalize the same
function so $C_{\text{align}} = 1$, and the multiplier is initialized uniformly at
$\mu_0 = 1$. Eqn.~\eqref{eqn:app_ul_factor_general} then reduces to
$k = |g_b|/\texttt{tol}$, as used in Eqn.~\eqref{eqn:ul_factor}. 
 
\subsection{Initial LL learning rate selection}
\label{app:impl_lr_probe}
 
The learning rate at which the LL is stable depends strongly on the
parameterization. The stable range of the CNN parameterization, for instance,
differs from that of the standard parameterization often by orders of magnitude,
and additionally depends on the network width, depth and initialization scheme. A
fixed default therefore does not transfer between the parameterizations used in
this work, and we determine $\eta_0$ per problem with the following heuristic.

\paragraph{Acceptance criteria}
Let $f$ denote the LL objective and let
$\boldsymbol{p}_k = \boldsymbol{x}_{k+1} - \boldsymbol{x}_k$ be the update actually
applied by the LL optimizer at step $k$. A candidate learning rate $\eta$ must
satisfy two tests, applied at different scopes.

The first is a per-step sufficient-decrease test. At every step
$k = 0, \dots, t_{\text{start}}-1$ we require
\begin{equation}
    f(\boldsymbol{x}_{k+1}) - f(\boldsymbol{x}_{k}) \;\leq\;
    \tfrac{1}{2}\, \nabla f(\boldsymbol{x}_k)^{\top} \boldsymbol{p}_k,
    \label{eqn:probe_c1}
\end{equation}
\noindent i.e. the realized decrease must be at least half the decrease predicted
by the first-order model along the step actually taken. The trial is terminated at
the first step that violates Eqn.~\eqref{eqn:probe_c1}. The second test is applied once, to the trial as a whole, on the cumulative
displacement:
\begin{equation}
    \frac{\bigl|f(\boldsymbol{x}_{t_{\text{start}}}) - f(\boldsymbol{x}_0)\bigr|}
         {\bigl|f(\boldsymbol{x}_0)\bigr|} \;\leq\; \varepsilon_{\text{rel}}.
    \label{eqn:probe_c2}
\end{equation}
with $\varepsilon_{\text{rel}}=0.1\%$.
A candidate is accepted if it completes all $t_{\text{start}}$ steps under
Eqn.~\eqref{eqn:probe_c1} and then satisfies Eqn.~\eqref{eqn:probe_c2}. The two are
complementary---Eqn.~\eqref{eqn:probe_c1} rejects learning rates that overshoot
within a step, Eqn.~\eqref{eqn:probe_c2} rejects those that make well-behaved
steps but travel too far overall---and together they select the largest learning
rate that is both productive and conservative.


\paragraph{Search.}
We place $N$ candidates on a geometric grid spanning
$[10^{-8}, 10^{0}]$ and scan upward from the lower bound, evaluating
Eqns.~\eqref{eqn:probe_c1}--\eqref{eqn:probe_c2} at each candidate and stopping at
the first failure. Scanning in this direction and stopping at the first failure
ensures that the accepted value lies in the stable region contiguous with the
smallest learning rates rather than in an isolated pocket higher up the grid. The
last passing candidate $\eta^{-}$ and the first failing candidate $\eta^{+}$
bracket the stability boundary. We then bisect in $\log \eta$, maintaining the
invariant that $\eta^{-}$ passes and $\eta^{+}$ fails, for $8$ steps. The
resulting boundary estimate $\eta^{\star}$ is the final passing endpoint, and we
initialize
\begin{equation}
    \eta_0 = 0.5\,\eta^{\star},
\end{equation}
backing off by a factor of two so that the bilevel optimization does not begin at
the edge of the stable region.

\paragraph{Cost.}
The scan uses at most $20$ trials and the bisection at most $8$, so the probe
requires at most $28$ trials of $t_{\text{start}}$ LL steps each. Every step
requires one forward evaluation of the LL objective and one LL gradient
evaluation; no hypergradient, no unrolling and no UL evaluation is involved, and
the optimizer used in the probe is exactly the one used subsequently in the
bilevel optimization (but with its state reset).

\subsection{Pseudo-first-order approximation}

The pseudo-first-order approximation of the hypergradient reduces to 
a single line in JAX: when computing the LL gradient, 
\texttt{jax.lax.stop\_gradient} is applied to the design variables 
before the gradient call,
\pyinline{jax.value_and_grad(f_LL)(jax.lax.stop_gradient(x), y)}
This prevents JAX from differentiating through the dependence of 
$\boldsymbol{x}$ on $\boldsymbol{y}$ across LL iterations, setting 
$\mathbf{A}_j = \mathbf{I}$ in the reverse-mode recursion and eliminating the Hessian term from the hypergradient computation.

\section{Analytical SMD problems}
\label{app:smd_details}

For the results presented in Fig.~\ref{fig:smd_part1} and 
Fig.~\ref{fig:smd_part2}, the LL dimensionality is set to two. The 
analytical LL optimum is $\boldsymbol{x}_\star = [0,\, \arctan(y_2)]$. 
The LL optimizer is Adam with learning rate $10^{-2}$ and gradient 
clipping threshold $g_c = 1.0$; the UL optimizer is Adam with learning 
rate $10^{-1}$, also with clipped gradients ($g_c = 1.0$).

The histogram in Fig.~\hyperref[fig:smd_part2]{\ref*{fig:smd_part2}b} 
is constructed as follows. For each bilevel optimization run and each 
annealing stage $t$, the number of UL iterations $k$ performed at that 
stage is accumulated across all runs. The resulting counts are then 
normalized by the total number of UL iterations across all runs and 
stages, so that the histogram reflects the fraction of the total 
optimization budget spent at each value of $t$.

\subsection{Warmstarting and non-convexity at the lower-level}
\label{sec:ncr}
\begin{figure*}[t]
    \centering
    \includegraphics[width=1\textwidth]{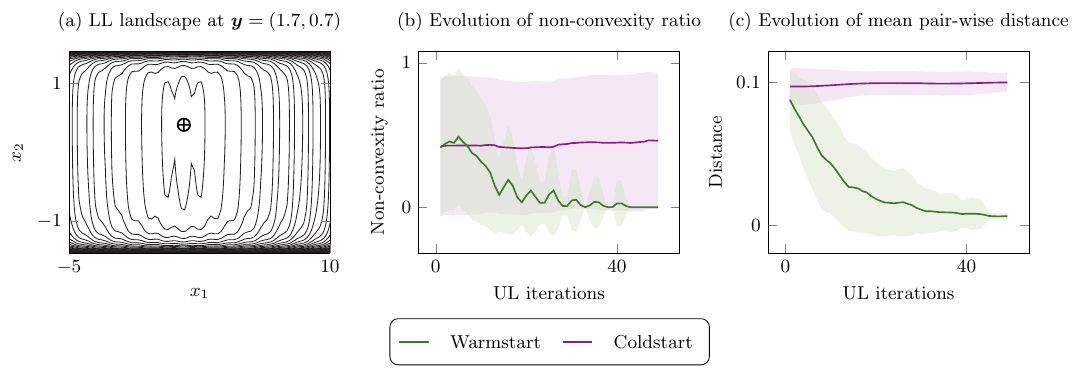}
\caption{Non-convexity exposure of the LL trajectory in the modified 
bilevel problem of Eqn.~\eqref{eq:smd3a}. Several bilevel optimizations 
are run from randomly sampled UL--LL initializations with a fixed LL 
horizon $t = 40$, differing only in the LL initialization scheme 
(coldstart or warmstart). At each UL iteration $k$, pairs of points 
are sampled from the LL trajectory and the non-convexity ratio 
(Eqn.~\eqref{eq:ncr}) is computed; results are aggregated across all 
runs. (a) An example LL landscape showing multiple local minima. 
(b) Non-convexity ratio versus UL iteration $k$ for coldstart and 
warmstart; shaded regions show one standard deviation across runs. 
(c) Mean pairwise distance within the LL trajectory at each UL 
iteration.}
    \label{fig:warm_ncr}
\end{figure*}

To examine why warmstarting yields smoother bilevel optimization than 
coldstarting, we measure the non-convexity of the LL landscape along 
the trajectory traversed by the optimizer under each initialization 
scheme. We construct a 
modified bilevel test problem with a 2D convex UL objective and a 2D
multimodal LL whose basin centers shift with the UL variables:

\begin{equation}
\label{eq:smd3a}
\begin{aligned}
F(\bs{y}, \bs{x}_\star) 
&= y_1^2 + x_{1\star}^2 
   + y_2^2  + \left(y_2^2 - \tan x_{2\star}\right)^2, \\
f(\bs{y}, \bs{x}) 
&= y_1^2 + 1 + (x_1 - y_1)^2 - \cos 2\pi (x_1 - y_1)  \\
&\qquad + \left(y_2^2 - \tan x_{2}\right)^2,
\end{aligned}
\end{equation}
\noindent where $\bs{y} = (y_1, y_2)$ and $\bs{x} = (x_1, x_2)$ are the UL and LL variables respectively. The Rastrigin-style term makes the LL multimodal, with 
basin centers tracking $y_1$ so that the basin structure is 
genuinely UL-dependent; the global LL minimum lies at 
$\bs{x}_\star = (y_1, \arctan(y_2^2))$. The 
resulting UL objective is convex in $\bs{y}$, similar to SMD-1 but the LL landscape, shown in 
Fig.~\hyperref[fig:warm_ncr]{\ref*{fig:warm_ncr}a}, exhibits clearly separated local minima.

To quantify the non-convexity of the LL regions traversed at each UL 
iteration $k$, we use a midpoint-defect criterion~\cite{Tamura2019}. 
For any two LL points $\bs{a}, \bs{b}$ and UL point $\bs{y}^{(k)}$,
\begin{equation}
\begin{aligned}
\Delta(\bs{a}, \bs{b};\, \bs{y}^{(k)}) &= 
   f\!\left(\bs{y}^{(k)},\, \tfrac{\bs{a}+\bs{b}}{2}\right) \\ 
   &\qquad- \tfrac{1}{2}\bigl(f(\bs{y}^{(k)}, \bs{a}) 
   + f(\bs{y}^{(k)}, \bs{b})\bigr),
\end{aligned}
\end{equation}
\noindent with $\Delta > 0$ indicating a non-convex barrier between $\bs{a}$ 
and $\bs{b}$. The non-convexity ratio at iteration $k$ is the 
fraction of trajectory pairs for which this defect is positive:
\begin{equation}\label{eq:ncr}
\mathrm{NCR}(k) = \frac{1}{|\mathcal{P}_k|}    \sum_{(\bs{a},\bs{b})\,\in\,\mathcal{P}_k} 
   \mathbf{1}\!\left[\Delta(\bs{a}, \bs{b};\, \bs{y}^{(k)}) > 0\right],
\end{equation}
where $\mathcal{P}_k$ is the set of unordered pairs of distinct points 
along the LL trajectory at iteration $k$, subsampled to at most $1000$ 
pairs when the trajectory length exceeds this limit. A value near zero 
indicates the trajectory remains within a locally convex region; values 
near one indicate frequent crossing of non-convex barriers.
 
\paragraph*{Experimental setup.}
We run paired coldstart and warmstart optimizations from $75$ 
initializations sampled uniformly at random over both the UL and LL 
variable ranges. Both levels use Adam with gradient clipping 
($g_c = 1.0$): the LL optimizer uses learning rate $10^{-2}$ and the 
UL optimizer uses learning rate $10^{-2}$. The pseudo-first-order 
approximation and gradient clipping are applied identically in both 
regimes. No annealing is performed; each run uses a fixed LL horizon $t = 40$ 
and performs $50$ UL iterations.\footnote{Results for $t = 20$ follow 
the same trend and are omitted for brevity.} The NCR is computed from the LL trajectory 
at each UL iteration, with pairs subsampled to at most $1000$ per 
trajectory.

Fig.~\hyperref[fig:warm_ncr]{\ref*{fig:warm_ncr}b} shows that 
warmstarting consistently produces lower NCR than coldstarting across 
all UL iterations and both values of $t$, confirming that warmstarted 
trajectories remain within locally convex neighborhoods of the LL 
optimum. Fig.~\hyperref[fig:warm_ncr]{\ref*{fig:warm_ncr}c} shows that 
warmstarted trajectories are also substantially shorter in mean pairwise 
distance, reflecting that the LL solve is initialized near the previous 
optimum and therefore traverses only a small neighborhood. Combined with 
the LL residual results in the main text 
(Fig.~\hyperref[fig:smd_part2]{\ref*{fig:smd_part2}d}), these results 
establish that warmstarting's contribution to UL smoothness operates 
through LL convergence: by keeping the LL trajectory close to the 
optimum, warmstarting confines it to a locally convex region and yields 
well-behaved hypergradients.

\section{Compliance minimization: experimental details}\label{app:comp}

We use the standard MBB beam boundary conditions~\cite{andreassen2011efficient} 
with a unit load applied at the top-left node. The mesh resolution is 
$N_x \times N_y = 144 \times 48$, chosen to be divisible by eight for 
compatibility with the CNN architecture of~\cite{hoyer}. The target 
volume fraction is $30\%$, with material properties $E = 1$, 
$E_{\min} = 10^{-6}$, and $\nu = 0.3$. We use bilinear quadrilateral 
(Q4) elements under plane stress and solve the equilibrium equations 
with a direct solver. For the 2D landscape visualization, $r_{\min} = 
2.0$ and $p = 3.0$ are held fixed. For the four-hyperparameter tuning 
experiment, bounds are given in Tab.~\ref{tab:app_hp_lim} and the 
learning rate is parameterized in log space. The initial LL learning 
rate is set to $\eta^{(0)} \approx 1 \times 10^{-8}$ as selected by the 
probe procedure described in Appendix~\ref{app:impl_lr_probe}. To normalize the UL loss, reference values were calculated based on a random network initialization and kept fixed across all runs. Consequently, the UL objective is independent of the run settings, enabling direct comparisons between BO and BOTH. All 
other BOTH settings follow Appendix~\ref{app:hyperto_details}.

\begin{table}[h!]
\centering
\caption{Bounds on UL variables for hyperparameter tuning experiment.}
\label{tab:app_hp_lim}
\begin{tabular}{@{}lll@{}}
\toprule
UL variable & $y_{\min}$ & $y_{\max}$ \\ \midrule
Learning rate $\eta$      & $10^{-8}$  & $10^{-1.5}$   \\
Heaviside projection $\beta$     & $0.05$     & $15.0$     \\
Filter radius $r_{\min}$  & $1.5$      & $10.0$     \\
SIMP penalty $p$         & $1.0$      & $10.0$     \\ \bottomrule
\end{tabular}%
\end{table}


\begin{figure}[t]
    \centering
    \includegraphics[width=1\columnwidth]{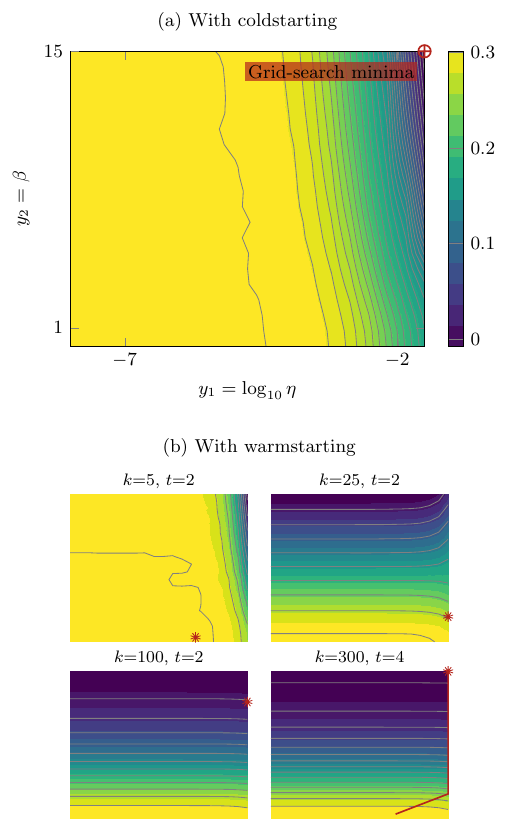}
\caption{The UL landscape for standard topology optimization. (a) Landscape obtained by independently solving the lower-level problem from the same fixed initialization for each hyperparameter pair $(\eta,\beta)$, corresponding to the landscape observed by BO. (b) Effective landscape encountered by BOTH under warmstarted optimization; the red asterisk marks the current UL iterate and the trailing 
line shows the trajectory followed. Here, $k$ denotes the UL iteration index and $t$ denotes the number of LL optimization steps performed at that iteration.}
\label{fig:app_comp2d_landscape}
\end{figure}


\begin{figure*}[t]
    \centering
    \includegraphics[width=1\textwidth]{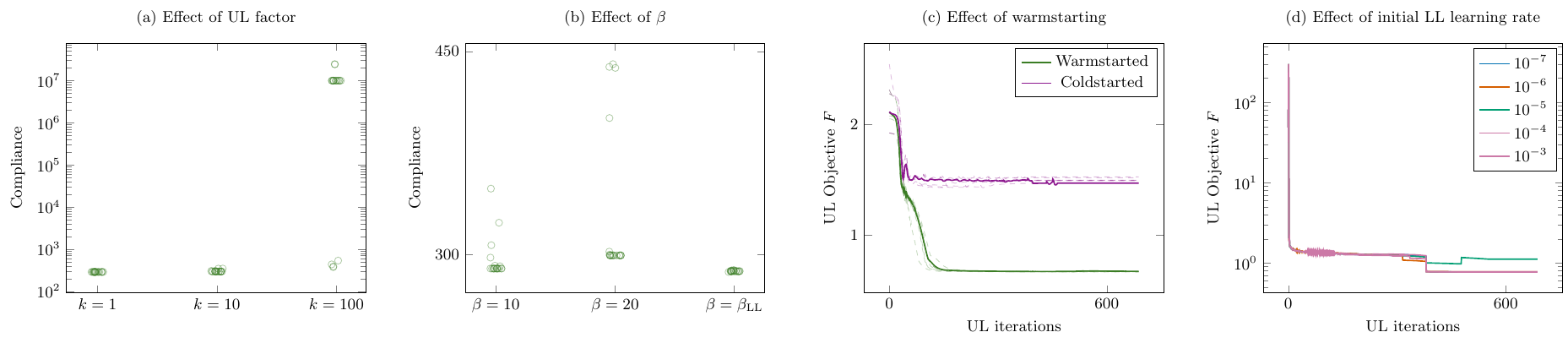}
\caption{Ablation study for compliance minimization with neural 
reparameterization, varying one component of the base configuration 
at a time across $6$ BOTH runs; run-to-run variation arises 
solely from differences in network or hyperparameter initialization. 
(a) Effect of the weighting coefficient $k$ in the UL objective, 
which controls the relative contribution of compliance and 
discreteness. (b) Effect of the Heaviside sharpness $\beta$ used in 
the UL objective; $\beta = \beta_{\text{LL}}$ indicates that the LL 
value of $\beta$ is reused at the UL, introducing a direct 
hypergradient term. (c) Effect of the initialization scheme: 
warmstarting versus coldstarting. (d) Effect of the initial LL 
learning rate $\eta^{(0)}$, normally selected by the probe procedure 
but set manually here.}
\label{fig:comp_ablation}
\end{figure*}

\subsection{Network architecture}

We use the CNN architecture from~\cite{hoyer}, which is the decoder 
of a U-Net. The network receives a trainable latent vector of size 
$128$ as input, passed through a fully connected layer with $\tanh$ 
activation to produce a tensor of shape $(N_x/8,\, N_y/8,\, 32)$. 
This tensor is then processed through five successive hidden 
layers: each layer applies bilinear interpolation to double the 
spatial resolution, followed by layer normalization (zero mean, unit 
variance), a convolution, addition of a per-pixel trainable bias, and 
a $\tanh$ activation. This is repeated until the output matches the 
target mesh resolution $(N_x, N_y)$. We refer the reader 
to~\cite{hoyer} and the associated code for full architectural 
details.

\subsection{Additional results and ablation study}

Fig.~\ref{fig:app_comp2d_landscape} shows the result of applying BOTH to 
the standard density parameterization, i.e., without a neural 
network. To use Adam, the sigmoidal projection is applied at each LL 
iteration to enforce the volume constraint. For this parameterization, the initial 
density field is fixed at the uniform value $(\boldsymbol{x}_0^{(0)})_e = 
V_0/V = 0.3$, as per standard practice. We tune the learning rate 
$\eta$ and Heaviside sharpness $\beta$, holding the SIMP penalty 
exponent and filter radius fixed at $p = 3.0$ and $r_{\min} = 2.0$ 
respectively. The initial LL learning rate is determined by the probe 
procedure and is approximately $\eta^{(0)} \approx 10^{-2.5}$. The 
coldstart UL landscape is considerably smoother than in the neural 
reparameterization case.

Additionally, we report the sensitivity of BOTH to four user choices: the 
weighting coefficient $k$ in Eqn.~\eqref{eq:compliance_UL}, the 
Heaviside sharpness $\beta$ used to evaluate $F$,  the 
LL initialization scheme, and the initial LL learning rate. The configuration used in the main text (Fig.~\ref{fig:comp4d}) 
serves as the base: $k = 1$, $\beta = \beta_{\text{LL}}$ at the UL, and 
warmstarting. Each ablation changes one component at a time. Results are shown in Fig.~\ref{fig:comp_ablation}.

Fig.~\hyperref[fig:comp_ablation]{\ref*{fig:comp_ablation}a} shows 
the effect of varying $k$. Increasing $k$ amplifies the weight of the 
discreteness term relative to compliance; since both terms are 
normalized at initialization, larger $k$ causes the discreteness 
objective to dominate, producing black-and-white but structurally 
disconnected designs for a few of the runs.

Fig.~\hyperref[fig:comp_ablation]{\ref*{fig:comp_ablation}b} shows 
the effect of the $\beta$ value used in the UL objective. To ensure a 
fair comparison, all final designs are projected at $\beta = 64$ 
before evaluation. BOTH is more robust to this choice; the 
worst results occur at $\beta = 20$, likely due to the reduced 
hypergradient signal and increased nonlinearity at high projection 
sharpness. Notably, using the same $\beta$ as the LL (which 
introduces a direct hypergradient term at the UL) yields the best 
results, suggesting that the direct term provides a useful additional 
gradient signal.

Fig.~\hyperref[fig:comp_ablation]{\ref*{fig:comp_ablation}c} 
confirms the importance of warmstarting: all cold-started runs 
perform substantially worse than their warmstarted counterparts. 
Fig.~\hyperref[fig:comp_ablation]{\ref*{fig:comp_ablation}d} shows 
that BOTH is robust to the choice of initial learning rate across 
the range tested.

\subsection{Bayesian optimization baseline}

We compare BOTH against BO as a strong baseline for 
low-dimensional hyperparameter optimization. BO is implemented using 
the Optuna package~\cite{optuna_2019} with a Gaussian process (GP) 
surrogate, a Mat\'{e}rn kernel, and expected improvement as the 
acquisition function (with default kernel hyperparameters as set by Optuna). Each BO run uses $5$ startup trials sampled 
uniformly from the hyperparameter bounds before GP-guided acquisition 
begins; we tested $5$, $10$, and $15$ startup trials and found $5$ 
to give the strongest final performance under a matched computational 
budget.

For the neural reparameterization (CNN) case, we run $8$ independent 
BO trials per network initialization (three initializations), giving 
$24$ BO runs in total; trial independence is enforced by using 
distinct random seeds for the GP. For the standard density parameterization case (tuning only two hyperparameters), run-to-run 
variability is introduced by sweeping the SIMP penalty $p$ over 
$\{2.5, 3.0, 4.0\}$ paired with the same $8$ random GP seeds used 
in the CNN case, yielding $24$ runs in total.

The two methods are compared on the basis of the number of linear solves used. Each BOTH UL update requires $4t + 2$ linear solves, 
whereas each BO sample requires $2t + 1$ solves: $2t$ to run TO for 
$t$ LL iterations plus one additional solve to evaluate the UL 
objective. Unlike BOTH, each BO sample runs TO to convergence 
before contributing a single data point. Convergence within each BO sample is declared when the LL objective 
does not improve by more than $0.1\%$ over the last $5$ iterations, 
with a maximum of $t = 150$ LL iterations per sample; only the 
iterations actually used count toward the budget. The same UL 
objective $F$ and hyperparameter bounds (Tab.~\ref{tab:app_hp_lim}) 
are used for both methods.

\section{Two-dimensional stress example}\label{app:stress2d}

\subsection{Quadratic Penalty Method and Epsilon Relaxation}
\label{app:qpm_eps_relax}
This residual is the 
\emph{tightest} relaxation achievable at a given $\mu^{(k)}$; 
approximate (finite-$t$) minimization yields larger violations. 
A constrained optimization problem of the form:
\begin{equation}
    \min_{\bs{x}} f(\bs{x}) \quad \text{s.t.} \quad g(\bs{x}) = 0
\end{equation}
can be addressed using the quadratic penalty method (QPM), which solves a sequence of unconstrained subproblems with augmented objective:
\begin{equation}
    \min_{\bs{x}} Q_k(\bs{x}) = f(\bs{x}) + \frac{\mu_k}{2}g(\bs{x})^2,
\end{equation}
where $\mu_k > 0$ is an increasing penalty parameter. Let $\bs{x}^\star$ denote the minimizer of $Q_k$. The first-order optimality condition is
\begin{equation}
    \nabla_{\bs{x}} Q_k(\bs{x}^\star) = \nabla f(\bs{x}^\star) + \mu_k g(\bs{x}^\star) \nabla g(\bs{x}^\star) = 0. \label{eq:qpm_kkt}
\end{equation}

Now consider the $\epsilon$-relaxed version of the original problem, where the equality constraint is perturbed to $g(\bs{x}) = \epsilon$. The Lagrangian is $L(\bs{x}, y) = f(\bs{x}) + y\bigl(g(\bs{x}) - \epsilon\bigr)$, where $y$ is the Lagrange multiplier. The KKT conditions for the relaxed problem at its optimal point $\bs{x}^\star$ are:
\begin{align}
    \nabla_{\bs{x}} L(\bs{x}^\star, y) &= \nabla f(\bs{x}^\star) + y \nabla g(\bs{x}^\star) = 0, \label{eq:eps_stat} \\
    g(\bs{x}^\star) &= \epsilon. \label{eq:eps_constraint}
\end{align}

Comparing \eqref{eq:qpm_kkt} and \eqref{eq:eps_stat}, the two stationarity conditions are structurally identical provided we identify
\begin{equation}
    y = \mu_k \, g(\bs{x}^\star).
\end{equation}
Substituting the relaxed constraint \eqref{eq:eps_constraint} then gives $y = \mu_k \epsilon$. Thus, the minimizer of the $k$-th QPM subproblem is equivalently the solution to the $\epsilon$-relaxed constraint problem with $\epsilon = g(\bs{x}^\star)$ and optimal Lagrange multiplier $y = \mu_k \epsilon$. As $\mu_k \to \infty$, $\epsilon \to 0$ and the relaxed problem recovers the original equality constraint.

This result extends naturally to inequality constraints $g(\bs{x}) \leq 0$, where active constraints are handled analogously and inactive constraints do not contribute to the penalty.

\section{L-shaped beam design example}
The stress tensor at the element level is first calculated using $\hat{\bs{\sigma}}_e = C B \bs{u}_e$, where $C$ is the constitutive matrix assuming $E=1$ and $\nu=0.3$. The von mises estimate is calculated as $\sigma_v = \sqrt{\hat{\bs{\sigma}}^\top.T.\hat{\bs{\sigma}}}$` and is the stress measure i.e. $|\sigma_i| = \sigma_v$. We pad the domain according to \cite{amir_stress}, where all free boundaries except the support locations have void padding. The helps eliminate edge effects of filtering by treating internal and external edges consistently. Loading is made distributed and non-uniform over fixed passive material patch that extends into padding so that filtering does not cause gray values at loading locations.

\label{app:lshape}
\bibliographystyle{unsrt}  
\bibliography{sn-bibliography}

\end{document}